\documentclass[times, review, 10pt]{elsarticle}

\usepackage{amssymb}
\usepackage{amsmath}
\usepackage{url}
\usepackage{booktabs}       
\usepackage{amsfonts}       
\usepackage{nicefrac}       
\usepackage{microtype}      
\usepackage{xcolor}         
\usepackage{graphicx}
\usepackage{mathtools,xspace}
\usepackage{multirow}
\usepackage{colortbl}
\usepackage{wrapfig}
\usepackage{longtable}
\usepackage{pifont}
\usepackage{marvosym}
\usepackage{gensymb}

\definecolor{MyCiteColor}{rgb}{0.5, 0.1, 0.8}
\usepackage[colorlinks,citecolor=MyCiteColor]{hyperref}

\definecolor{myred}{rgb}{0.7, 0, 0}
\definecolor{myblue}{rgb}{0, 0.6, 1}
\definecolor{mblue}{rgb}{0, 0, 0.6}
\definecolor{lightblue}{rgb}{0.66,0.85,0.98}
\definecolor{violet}{rgb}{0.5, 0.1, 0.5}
\definecolor{orrange}{rgb}{0.95,0.55,0.35}
\definecolor{myred}{rgb}{0.95,0.65,0.65}
\newcommand{\cmark}{\color{myblue}{\ding{51}}}%
\newcommand{\xmark}{\color{myred}{\ding{55}}}%
\newcommand{\purplecmark}{\color{MyCiteColor}{\ding{51}}}%
\newcommand{\cxmark}{\textcolor{black}{\purplecmark\kern-1.1ex\raisebox{.7ex}{\rotatebox[origin=c]{125}{--}}\color{black}}}

\newcommand{\onedot}{\ifx\@let@token.\else.\null\fi\xspace}
\newcommand{\eg}{\emph{e.g}\onedot}
\newcommand{\ie}{\emph{i.e}\onedot}
\newcommand{\etc}{\emph{etc}} 

\usepackage[most]{tcolorbox}
\newtcolorbox[list inside=promptbox, auto counter, number within=section]{promptbox}[2][Prompt]{
    colback=black!5!white,
    colframe=#2,
    title=#1,
    fontupper=\small,
    breakable,
    enhanced,
    left=0pt,
    right=0pt,
    top=0pt,
    bottom=0pt,
    arc=5pt,
    boxrule=1pt,
}

\journal{Pattern Recognition}

\begin{document}

\begin{frontmatter}



\title{LogicOCR: Do Your Large Multimodal Models Excel at Logical Reasoning on Text-Rich Images?}


\author{
Maoyuan Ye$^{a,*}$, Haibin He$^{a,*}$, Qihuang Zhong$^a$, Jing Zhang$^{a,\dag}$, \\Juhua Liu$^{a,\dag}$, Bo Du$^a$\\
\{yemaoyuan, haibinhe, zhongqihuang, liujuhua, dubo\}@whu.edu.cn, jingzhang.cv@gmail.com\\
*: Equal contribution \quad \dag: Corresponding author
}

\affiliation{organization={School of Computer Science, National Engineering Research Center for Multimedia
Software, Institute of Artificial Intelligence, and Hubei Key Laboratory of Multimedia
and Network Communication Engineering, Wuhan University},
            city={Wuhan},
            state={Hubei},
            country={China}}

\begin{abstract}
Recent advances in Large Multimodal Models (LMMs) have revolutionized their reasoning and Optical Character Recognition (OCR) capabilities. However, their complex logical reasoning performance on text-rich images remains underexplored. 
To bridge this gap, we introduce \textbf{LogicOCR}, a benchmark comprising 2,780 questions with two subsets, \ie, LogicOCR-Gen with 1,100 multi-choice questions on generated images, and LogicOCR-Real with 1,680 meticulously designed free-form questions on real-world images.
For constructing LogicOCR-Gen, we first curate a text corpus from the Chinese National Civil Servant Examination, and customize an automatic pipeline to steer GPT-Image-1 to generate images with varied layouts and fonts, ensuring contextual relevance and visual realism. Then, the generated images are manually verified.
We evaluate a range of representative LMMs under Chain-of-Thought (CoT) and direct-answer settings. 
Our multi-dimensional analysis reveals key insights, such as the impact of test-time scaling, input modality differences, and sensitivity to visual-text orientation. Notably, LMMs still lag in multimodal reasoning compared to text-only inputs, indicating that they have not fully bridged visual reading with reasoning.
Moreover, we propose \textbf{TextCue}, a training-free method that enhances LMMs’ perception of image regions containing important text cues for solving questions. We leverage LMMs' attention maps and an off-the-shelf text segmentation specialist to determine the region, which is then cropped and enlarged to augment the original image. Experiments show its effectiveness, \eg, a 1.8\% accuracy gain over LLaVA-OV-1.5-8B under the CoT setting.
Our benchmark is available at \href{https://github.com/MiliLab/LogicOCR}{LogicOCR}.
\end{abstract}



\begin{keyword}


Multimodal Reasoning \sep Text-Rich Images \sep Benchmark \sep Training-Free Method
\end{keyword}

\end{frontmatter}


\section{Introduction}
\label{introduction}

Recent advances in Large Multimodal Models (LMMs)~\cite{gpt4o,qwen2_5vl,internvl3, kimivl} have revolutionized their reasoning and Optical Character Recognition (OCR) abilities. While these skills are crucial for real-world applications, LMMs' ability to perform complex logical reasoning on text-rich images remains underexplored. Table~\ref{tab:1} compares representative benchmarks for multimodal reasoning and OCR. Existing multimodal reasoning datasets~\cite{mmmu,zhang2025mathverse} often overlook samples with dense visual text and require extensive STEM (Science, Technology, Engineering, Mathematics) knowledge, making it difficult to isolate reasoning ability from subject expertise. In contrast, most OCR-related benchmarks~\cite{singh2019towards,mathew2021docvqa,liu2024ocrbench,ccocr} lack complexity, potentially overstating LMM progress in integrating reading, understanding, and reasoning. Although CharXiv~\cite{wangcharxiv} and OCRBench v2~\cite{ocrbenchv2} include reasoning subsets, these are either narrowly focused on chart interpretation or rely on STEM knowledge.

To address the gap, we introduce LogicOCR, a benchmark comprising 2,780 questions designed to assess LMMs' ability to perform complex reasoning on text-rich images with minimal reliance on STEM knowledge. LogicOCR consists of two subsets, \ie, LogicOCR-Gen with 1,100 multi-choice questions on generated images, and LogicOCR-Real with 1,680 free-form questions on real-world images.
We build LogicOCR-Gen from a curated corpus of pure logical reasoning questions from the National Civil Servant Examination of China, rendering them into diverse, realistic images. To achieve this, we develop an automated, scalable pipeline. Specifically, we customize prompts and steer GPT-Image-1~\cite{gptimage1} to generate text-rich images with varied layouts (\eg, interleaved text and illustrations, backgrounds), fonts (handwritten and standard), ensuring contextual relevance and visual realism. For background style layout, the background descriptions are generated using Qwen2.5-14B~\cite{qwen2.5} based on the question context. All generated images undergo manual quality control to remove low-quality samples. 
For LogicOCR-Real, we collect real-world images from existing datasets and publicly available web sources. Then, the questions in LogicOCR-Real are thoughtfully crafted by experts based on the textual cues from images.

\begin{table*}[t!]
\centering
\setlength{\tabcolsep}{16pt}
\caption{
Comparison of multimodal reasoning and OCR-related benchmarks. The test set sizes are reported, with only the reasoning subsets of CharXiv and OCRBench v2 included. `Knwl. Free' refers to STEM knowledge free.
}
\resizebox{1\linewidth}{!}{\begin{tabular}{lccccc}
\hline
\toprule
\small \multirow{2}{*}{\textbf{Benchmarks}} &\small \textbf{Data} &\small \textbf{Complex} &\small \textbf{Knwl.} &\small \textbf{Diverse} &\small \textbf{Image} \\
&\small \textbf{Size} &\small \textbf{Reasoning} &\small \textbf{Free} &\small \textbf{Topics} &\small \textbf{Type} \\
\midrule
\rowcolor{gray!15} \multicolumn{6}{c}{\textit{\small Multimodal Reasoning}} \\
MathVista~\cite{lumathvista} &6.1K &\cmark &\xmark &\cmark &\small Real \\
MMMU~\cite{mmmu} &10.5K &\cmark &\xmark &\cmark &\small Real \\
MATH-Vision~\cite{wang2024measuring} &3.0K &\cmark &\xmark &\cmark &\small Real \\
MathVerse~\cite{zhang2025mathverse} &2.6K &\cmark &\xmark &\cmark &\small Real \\

\rowcolor{gray!15} \multicolumn{6}{c}{\textit{\small OCR-Related}} \\
TextVQA~\cite{singh2019towards} &5.7K &\xmark &\cmark &\cmark &\small Real\\
DocVQA~\cite{mathew2021docvqa} &5.2K &\xmark &\cmark &\xmark &\small Real \\
ChartQA~\cite{masry2022chartqa} &2.5K &\xmark &\cmark &\xmark &\small Real \\
OCRBench~\cite{liu2024ocrbench} &1.0K &\xmark &\cmark &\cmark &\small Real \\
CC-OCR~\cite{ccocr} &7.1K &\xmark &\cmark &\cmark &\small Real \\

\rowcolor{gray!15} \multicolumn{6}{c}{\textit{\small Multimodal Reasoning \& OCR-Related}} \\
CharXiv~\cite{wangcharxiv} &1.3K &\cmark &\cmark &\xmark &\small Real \\
OCRBench v2~\cite{ocrbenchv2} &2.2K &\cmark &\xmark &\cmark &\small Real \\

\midrule
LogicOCR (ours) &2.7K &\cmark &\cmark &\cmark &\small Real + Generated \\

\bottomrule
\hline
\end{tabular}}
\label{tab:1}
\end{table*}

We evaluate a range of state-of-the-art LMMs, including open-source models like Qwen2.5-VL~\cite{qwen2_5vl} and InternVL3~\cite{internvl3}, as well as proprietary models such as Gemini-2.5-Pro~\cite{gemini25pro}, under both direct answering and Chain-of-Thought (CoT) settings. Through multi-dimensional analysis, we reveal several key findings, such as:
\ding{182} \textbf{CoT does not consistently improve accuracy}—most models fail to reason better step-by-step on LogicOCR-Gen, suggesting flaws in reasoning paths.
\ding{183} \textbf{Test-time scaling is effective in improving performance}, though its efficiency remains limited.
\ding{184} \textbf{Reasoning over text-rich images remains a bottleneck}, suggesting that LMMs still struggle to fully bridge visual reading with reasoning.
\ding{185} \textbf{OCR robustness is still a major weakness}—image rotation perturbation can reduce accuracy to near-random levels.

Moreover, we propose a training-free visual cropping method, \textbf{TextCue}, which enhances LMMs' perception of image region containing important cues. Specifically, within a range of Large Language Model (LLM) layers, we adaptively select the most salient relative attention map as the importance map for visual cropping. Next, a rough box region is determined based on the attention map. Subsequently, an off-the-shelf text segmentation model is used to refine the box region, which is then cropped, enlarged, and concatenated with the original image in the second forward pass.

In summary, our main contributions are three-fold:
1) We construct the LogicOCR benchmark which evaluates LMMs' logical reasoning on text-rich images with minimal reliance on STEM knowledge.
2) We propose a scalable, automated pipeline using GPT-Image-1 to convert text corpora into diverse and realistic images.
3) We conduct comprehensive evaluations of LMMs and highlight key insights to guide future improvements in multimodal reasoning. Moreover, we provide a training-free TextCue method to enhance the baseline LMM on LogicOCR, \eg, increasing 0.7\% and 1.8\% accuracy over LLaVA-OV-1.5-8B under direct answering and CoT setting.

The rest of this paper is organized as follows. In Section~\ref{related_word}, we briefly review the related works. In Section~\ref{LogicOCR}, we introduce the details of LogicOCR benchmark, which consists of LogicOCR-Gen and LogicOCR-Real. Section~\ref{TextCue} presents our proposed training-free method TextCue in detail. Section~\ref{experiments} reports and analyzes our experimental results. Lastly, we present the limitation and conclusion of our study in Section~\ref{Limitations} and Section~\ref{Conclusion}, respectively.
\section{Related Works}
\label{related_word}

\textbf{Multimodal Reasoning and OCR-Related Benchmarks}.
Several benchmarks have been developed to evaluate the multimodal reasoning abilities of LMMs. Some focus on both scientific knowledge and reasoning~\cite{mmmu,mmmu-pro}, while others emphasize mathematical, computational, or visual puzzle tasks~\cite{lumathvista,zhang2025mathverse,wang2024measuring,song2025visualpuzzles}. Most of them demand extensive knowledge on specific subjects but lack samples with dense visual-text content, limiting their assessment of reasoning over rich contextual information. In comparison, various OCR-related benchmarks~\cite{singh2019towards,liu2024ocrbench} have been introduced to assess LMMs' capabilities in text recognition, key information extraction, document parsing, and visual question answering. However, they often neglect reasoning complexity, and some, like DocVQA~\cite{mathew2021docvqa}, quickly reach performance saturation. Recent chart-focused benchmarks~\cite{wangcharxiv,masry2022chartqa} require numerical reasoning abilities but are limited to chart data. In contrast, our LogicOCR benchmark evaluates complex logical reasoning in diverse, text-rich images.

\textbf{Reasoning-Capable LMMs.}
LMMs have advanced from high-resolution image perception~\cite{liu2024textmonkey,mplug-docowl2,gpt4o} to increasingly sophisticated multimodal reasoning~\cite{qwen2_5vl,internvl3,qvq}. High-resolution input enables LMMs to read with human-like accuracy, while test-time scaling~\cite{o1} extends their capabilities to more complex tasks. However, a key question remains: have LMMs truly integrated visual reading and reasoning? This work also explores this question through our proposed LogicOCR benchmark.

\textbf{Visual Localization and Cropping for Enhancing LMMs.}
Recent researches~\cite{wu2024v,mllm_know} have recognized the limitations of LMMs in fine-grained visual perception, particularly when answering questions about small or visually subtle objects within complex scenes. Prior works have identified this challenge and proposed diverse strategies to mitigate it, ranging from high-resolution fine-tuning~\cite{qwen2_5vl} to multi-agent collaborative pipelines~\cite{wu2024v}. In contrast, ViCrop~\cite{mllm_know} offers a distinct perspective by demonstrating that LMMs often possess an implicit ability to localize the visual subject of a question within their internal representations. ViCrop demonstrates that the performance degradation on small objects stems from the failure of observation as opposed to localization. Building on these insights, ViCrop introduces a training-free method that extracts and leverages the internal representations of LMM to refine their visual perception. 
In this work, inspired by ViCrop, we build a training-free baseline method on LogicOCR. We further enable a more adaptive method to select the internal representations of LMM and incorporate an off-the-shelf text segmentation specialist to refine the visual cropping regions in document and natural scenes.
\section{LogicOCR Benchmark}
\label{LogicOCR}

LogicOCR comprises 2,780 questions, including multi-choice and free-form questions. It consists of two subsets, \ie, LogicOCR-Gen with 1,100 multi-choice questions on generated images, and LogicOCR-Real with 1,680 free-form questions on real-world images. Next, we introduce the data collection and construction process, statistics, and evaluation protocol in detail.

\subsection{Data Collection and Construction of LogicOCR-Gen}
\begin{figure*}[t!]
    \centering
    \includegraphics[width=\linewidth]{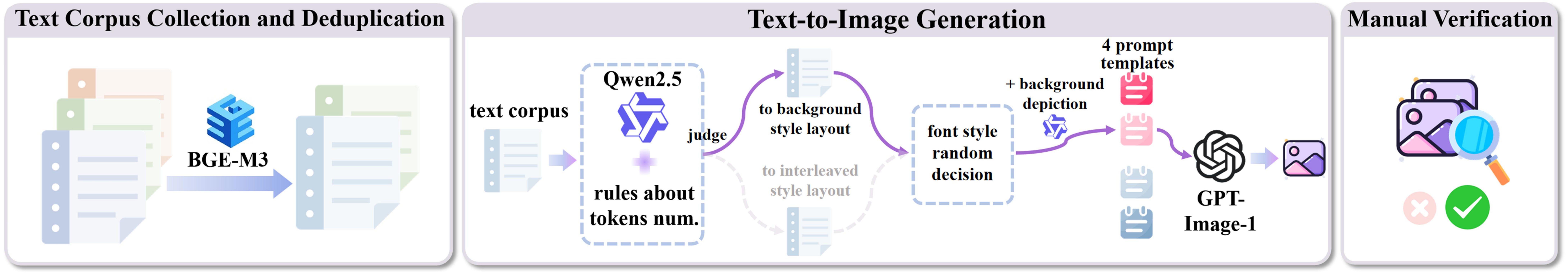}
    \includegraphics[width=\linewidth]{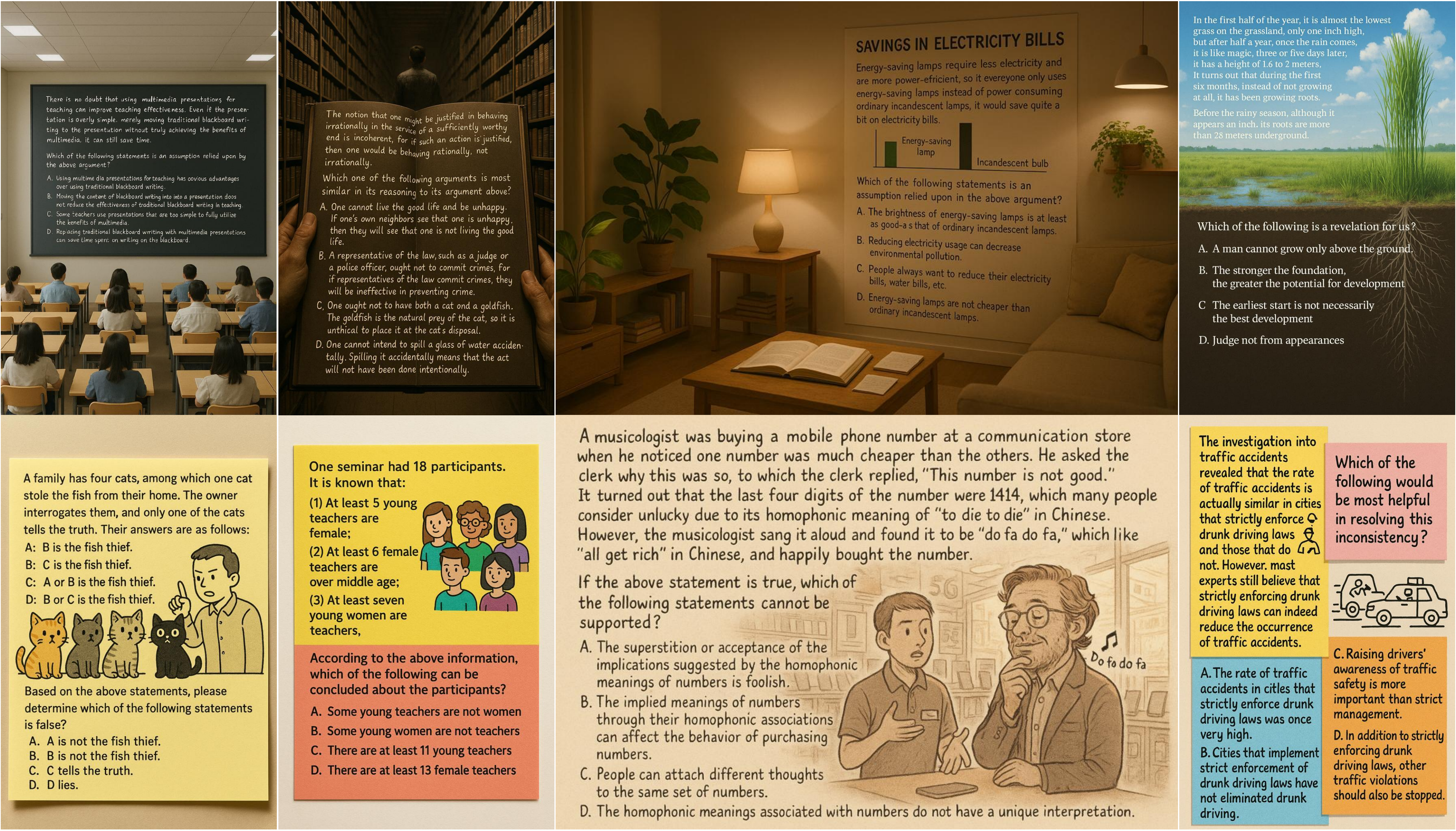}
    \caption{
    Illustration of the LogicOCR-Gen data construction process and sample images, showcasing background-style and text-illustration interleaved layouts from top to bottom.
    }
    \label{fig:main_figure}
    \vspace{-3mm}
\end{figure*}

To reduce the requirement of extensive STEM knowledge in reasoning questions and better utilize existing text sources for lower annotation costs, we collect a text corpus from the National Civil Servant Examination of China, which emphasizes pure logical reasoning. An automated pipeline is then designed to convert the corpus into diverse images, as shown in Fig.~\ref{fig:main_figure}.

\textbf{Text Corpus Collection and Deduplication.}
We reuse the text corpus from the test sets of LogiQA~\cite{logiqa} and LogiQA2.0~\cite{logiqa2.0}, both derived from the National Civil Servant Examination of China. For LogiQA samples containing mixed Chinese and English, we translate the original Chinese text into English. Each sample consists of three components: \textit{context}, \textit{question}, and \textit{options}, with the context providing all necessary information to answer the question. Since multiple questions may share the same context, we perform deduplication using the context, retaining only one sample per unique context. We generate embeddings for each context using BGE-M3~\cite{chen2024bge} and apply cosine similarity to identify and remove duplicates.

\textbf{Text-to-Image Generation.} 
We develop an automated pipeline that converts the text corpus into images using Qwen2.5~\cite{qwen2.5} and GPT-Image-1~\cite{gptimage1}. To enhance image diversity, generation success, and visual-text readability, we introduce four prompt templates that account for layout and font style when instructing GPT-Image-1.

Firstly, we design two layout styles: 1) full-frame background scenes with contrastive visual-text in the foreground, and 2) interleaved text and illustrations on randomly colored paper. To determine suitability for the background style, we use Qwen2.5-14B for evaluation and assign this layout to samples exceeding 250 tokens. The background scenes and illustrations are contextually aligned with the questions to enhance visual relevance and naturalness. To support this, we add tailored instructions into the prompt templates. For samples requiring background scenes, Qwen2.5-14B also generates short contextual descriptions, which are inserted into designated placeholders in the templates for text-to-image generation. For the interleaved style, GPT-Image-1 effectively produces suitable illustrations based on our prompt templates, so no additional image descriptions are needed.

Secondly, we define font style as either handwritten or non-handwritten, sampled randomly without specifying exact fonts. Combined with the two layout types, this yields four prompt templates, detailed in Appendix A. 
For GPT-Image-1 generation, we set the quality to ``high'' and the resolution to either $1536 \times 1024$ or $1024 \times 1536$. We demonstrate the high text rendering fidelity in Tab.~\ref{tab:4}.







\begin{wraptable}[18]{r}{5.5cm}
\centering
\setlength{\tabcolsep}{12pt}
\vspace{-0.35cm}
\caption{
LogicOCR-Gen statistics. 
Each sample may cover multiple reasoning categories.
}
\resizebox{5.5cm}{!}{
\begin{tabular}{lr}
\hline
\toprule
\textbf{Statistics} &\textbf{Value} \\
\midrule

\textbf{Questions}                     &1,100 \\
- LogiQA~\cite{logiqa} corpus                  &330 \\
- LogiQA2.0~\cite{logiqa2.0} corpus           &770 \\

\textbf{Reasoning Categories}                & \\
- categorical                                &50.5\%  \\
- sufficient conditional                     &86.3\%  \\
- necessary conditional                      &37.5\%  \\
- disjunctive                                &21.6\%  \\
- conjunctive                                &66.3\%  \\

\textbf{Answer Distribution}                & \\
\multicolumn{2}{l}{- A/B/C/D 24.0\%/23.3\%/27.7\%/25.0\%} \\

\midrule
\textbf{Image Characteristics}                & \\
- background style layout                 &41.5\% \\
- interleaved style layout                 &58.5\% \\
\arrayrulecolor{black!20}
\midrule
- non-handwritten style font                 &53.2\% \\
- handwritten style font                 &46.8\% \\
\midrule
- average words in image               &139.2 \\

\arrayrulecolor{black}
\bottomrule
\hline
\end{tabular} }
\label{tab:2}
\end{wraptable}

\textbf{Motivation of Generating Images Using GPT-Image-1.}
Since generated images and their source text are like modality twins, we can directly compare performance across input types by feeding plain-text questions and multimodal problems into the same LMM. This enables us to pinpoint some multimodal reasoning bottlenecks. GPT-Image-1 offers strong instruction-following and high-fidelity visual-text rendering, presenting a promising alternative to prior synthetic methods~\cite{gupta2016synthetic}. Unlike those approaches, it produces more realistic images without complex rule design. Notably, GPT-Image-1 can intelligently adapt text placement, color, and wrapping to fit surrounding elements.

\textbf{Manual Verification.}
Generated images are saved in JPEG format and resized to a maximum dimension of 1024 pixels to reduce the computation overhead of LMMs during evaluation. Each image undergoes a manual verification step to ensure text readability. Images lacking correct answer options or impairing judgment are discarded. The overall generation success rate is about 73\%.

\textbf{Statistics.}
The key statistics of LogicOCR-Gen are presented in Tab.~\ref{tab:2}. Based on the definition of corpus sources~\cite{logiqa,logiqa2.0}, five logical reasoning categories are considered: categorical, sufficient conditional, necessary conditional, disjunctive, and conjunctive. For LogiQA2.0~\cite{logiqa2.0}, we directly use the annotated reasoning type for each sample. Since LogiQA~\cite{logiqa} lacks such annotations, we classify its samples using o4-mini~\cite{o4mini} with few-shot examples. Note that one sample may correspond to multiple reasoning types.
Regarding image layout, 41.5\% follow a background-style format, while the rest adopt an interleaved layout. 
In addition, each image contains over 130 words.

\begin{figure*}
    \centering
    \includegraphics[width=\linewidth]{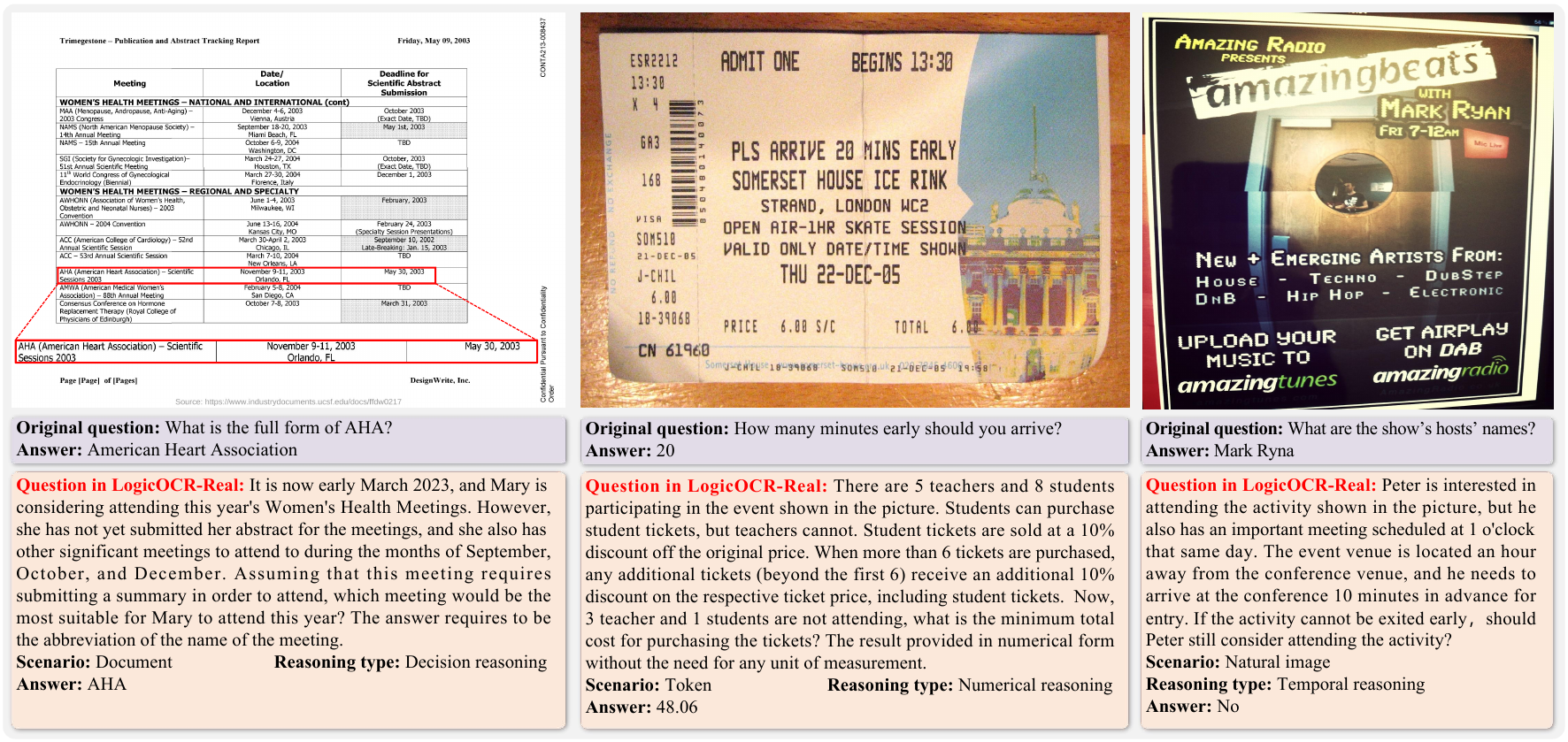}
    \caption{
    Comparison of the newly crafted multi-hop reasoning questions in LogicOCR-Real and the original questions in previous benchmarks.
    }
    \label{fig:logicocr_real}
    \vspace{-4mm}
\end{figure*}

\subsection{Data Collection and Construction of LogicOCR-Real}
Previous real-world benchmarks~\cite{mathew2021docvqa,singh2019towards,masry2022chartqa,dtvqa} often exhibit limited question complexity, rendering them insufficient for assessing the reasoning capabilities of contemporary LMMs, as shown in Fig.~\ref{fig:logicocr_real}. To bridge this gap, we construct the LogicOCR-Real subset, comprising 1,680 meticulously crafted free-form questions based on the textual cues from images.

\textbf{Image Collection.}
We collect the images from 1) publicly available web sources and 2) existing datasets, including DocVQA~\cite{mathew2021docvqa}, FUNSD~\cite{jaume2019funsd}, ChartQA~\cite{masry2022chartqa}, PlotQA~\cite{methani2020plotqa}, DT-VQA~\cite{dtvqa}, RICO~\cite{deka2017rico}, WildReceipt~\cite{wildreceipt}, \etc. 
These images cover a variety of scenarios and can be categorized into six types: 1) chart, 2) document (including printed documents, electronic documents, \etc), 3) screenshot (such as Graphical User Interface (GUI), excerpts from medical books, segments of product instructions and specifications), 4) product label, 5) natural image, and 6) token (including paper currency and tickets). 
We select the images which contain multiple cues for designing multi-hop reasoning questions.

\textbf{Question Design.}
The questions in LogicOCR-Real are thoughtfully designed and annotated by experts based on the textual cues from images, featuring higher complexity that requires multi-hop reasoning. 
For ease of evaluation, all questions are formulated as objective-type questions, with answers that are both concise and unambiguous. 
On average, each question consumes approximately \textit{half an hour}, including the selection of appropriate images, as well as the question design and annotation. 

\begin{wraptable}[16]{r}{5.5cm}
\centering
\setlength{\tabcolsep}{12pt}
\vspace{-0.8cm}
\caption{
LogicOCR-Real statistics.
}
\resizebox{5.5cm}{!}{
\begin{tabular}{lr}
\hline
\toprule
\textbf{Statistics} &\textbf{Value} \\
\midrule

\textbf{Questions}                     &1,680 \\
- newly designed questions                  &1,680 \\
- average question tokens                   &59.3 \\
- maximum question tokens                   &279 \\

\arrayrulecolor{black!20}
\midrule
\arrayrulecolor{black}
\textbf{Reasoning Categories}                & \\
- numerical reasoning                    &64.3\%  \\
- temporal reasoning                     &10.4\%  \\
- decision reasoning                     &6.2\%  \\
- conditional reasoning                  &19.1\%  \\

\midrule


\textbf{Scenarios}                & \\
- chart                                 &38.1\% \\
- document                              &28.0\% \\
- screenshot                            &14.0\% \\
- product label                         &8.8\% \\
- natural image                         &7.5\% \\
- token                                 &3.5\% \\

\arrayrulecolor{black}
\bottomrule
\hline
\end{tabular} }
\label{tab:statistics_real}
\end{wraptable}

Finally, these samples can be categorized into six reasoning classes, including 1) \textbf{numerical reasoning}, which involves data comparison analysis, statistical analysis, quantitative calculation, \etc, 2) \textbf{temporal reasoning}, which focuses on time relations such as sequence or duration, 3) \textbf{decision reasoning}, which requires making choices or judgments, and 4) \textbf{conditional reasoning}, which depends on logical inference based on the given context, and also includes other cases that do not fall into the above three categories.
Notably, certain reasoning categories are highly applicable to the domain of agents, particularly in areas such as decision reasoning. This highlights the benchmark’s versatility and its potential to drive advancements in these critical areas of research and application.

\textbf{Statistics.}
The key statistics are presented in Tab.~\ref{tab:statistics_real}, including the distribution of reasoning categories and image scenarios. Note that all questions are newly designed, encompassing 59.3 average question tokens and 279 maximum question tokens (calculated with the tokenizer of Qwen2.5-VL-7B).

\subsection{Evaluation Protocol}
Accuracy is used as the evaluation metric. Each LMM completion is parsed to extract the selected option letter for multi-choice question or the final answer for open-form question, then a binary score is assigned based on the ground-truth. For open-form questions, responses with the same meaning but different forms can also be considered correct. Hence, following previous works~\cite{he2024cmmu,pu2025judge}, we employ an external large model to serve as the judge for open-form questions, where GPT-4o-mini~\cite{gpt4o} is used.

\textbf{Prompt Templates for Evaluating LMMs.}
We evaluate LMMs under both CoT and direct answering settings, with prompts incorporating images and instructions. 
The templates used are detailed in Appendix B, including those for text-only input evaluation.
\begin{figure*}[t!]
\centering
    \includegraphics[width=1\textwidth]{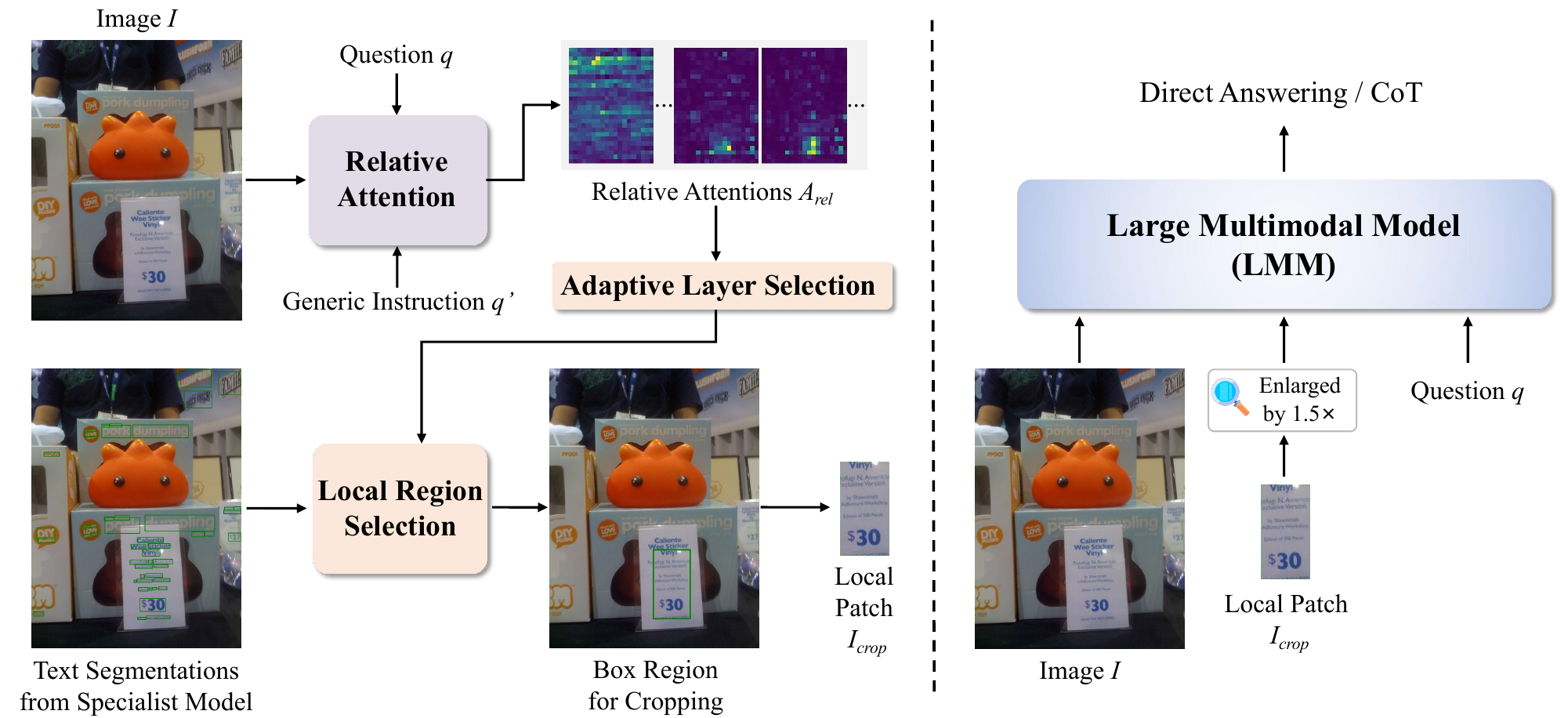}
    \caption{
    Illustration of TextCue, with visual cropping (left) and answering (right) stages. First, a relative attention mechanism is used to achieve relative attention maps. We introduce an adaptive layer selection scheme to determine the most salient attention map. Then, we convert the selected attention map into a rough box region, which is subsequently refined with the predictions of an off-the-shelf text segmentation model. Given the box region, a local patch is cropped, enlarged, and finally combined with the original image and question for answering.
    }
    \label{fig:textcue_method}
    
\end{figure*}

\section{A Simple Training-Free Visual Cropping Method: TextCue}
\label{TextCue}

Previous research~\cite{mllm_know} reveals that \textit{LMMs know where to look even when they answer incorrectly, and visual cropping can mitigate LMMs' perception limitation}.
Inspired by these findings, on LogicOCR, we provide a training-free visual-cropping method, TextCue. 
To be specific, during an additional pre-filling stage, we leverage a relative attention mechanism~\cite{mllm_know} to achieve relative attention maps within a range of language model layers. We introduce an adaptive layer selection (ALS) scheme to determine the most salient attention map. Next, based on the selected attention map, a rough box region is localized, which is subsequently refined with the prediction of an off-the-shelf text segmentation (TS) model. Finally, the box region is cropped, enlarged, and concatenated with the original image in the auto-regressive generation procedure (direct answering or using chain-of-thought). Overall, compared to the baseline method ViCrop~\cite{mllm_know}, our major modifications lie in the ALS scheme and the combination with a TS model.

\textbf{Relative Attention and Adaptive Layer Selection.}
Given an image-question pair ($I$, $q$), to measure the importance of each image token to the LMM's answer, the softmax attentions $A \in \mathbb{R}^{L \times 1 \times T}$ (averaged by the attention-head dimension) of the starting answer token to all image tokens through all layers of LLM can be used, where $L$ represent the number of LLM's layer while $T$ is the image token length fed into LLM. Since recent researches~\cite{kangsee} observe the attention sink phenomenon that LMMs tend to allocate high attention weights to specific image tokens, even when they are uninformative for answering the question. To distinguish the image tokens semantically relevant to question $q$, we adopt the relative attention mechanism which normalizes attentions with a generic instruction. Concretely, a fixed generic instruction $q'=$``Write a general description of the image.'' is used to calculate the relative attentions $A_{rel} = \frac{A(I, q)}{A(I, q')}$ under element-wise division. 

Given $A_{rel}$, through a preliminary empirical study, ViCrop selects the attention map $A_{rel}^m \in \mathbb{R}^{1 \times T}$ at the fixed layer $m$ as the importance map for subsequent local region selection. 
However, for reasoning questions on text-rich images, attentions may vary among different layers (see Appendix H). 
Hence, we further introduce the adaptive layer selection scheme. Specifically, we select $n$ attention maps $A_{rel}^{m:m+n-1} \in \mathbb{R}^{n \times T}$ and calculate the divergence between maximum and average value for each attention map: $div = \text{max}(A_{rel}^{m:m+n-1}, \text{dim=1}) - \text{average}(A_{rel}^{m:m+n-1}, \text{dim=1})$. The most salient attention map $A_{rel}^{m+\text{argmax}(div)}$ with the highest divergence is selected for local region selection.

\textbf{Local Region Selection for Visual Cropping.}
The attention map $A_{rel}^{m+\text{argmax}(div)}$ is first reshaped into a 2D shape, with each grid representing an image patch. To convert the attention map into a bounding box for visual cropping, the strategy of sliding windows with different sizes and aspect ratios is used. Following ViCrop, we define a set of windows with heights equal to a multiple ($\{1, 1.2, ..., 2\}$) of $224$ and varied aspect ratios ($\{1, 1.5, ..., 4\}$). 
Each window is slid across the image and its optimal position, where the sum of the importance map within the window is maximized, is identified. Following the selection of the best position for each window, the window exhibiting the greatest deviation between its internal sum and the average internal sum of its neighboring positions is chosen.

To further refine the window region, we integrate the predictions of a state-of-the-art text segmentation specialist, Hi-SAM~\cite{hisam}. For each image, we compute the intersection-over-union (IoU) scores between the window region and Hi-SAM's predictions. We collect the word bounding boxes with positive IoU and determine a minimum bounding rectangle as the final box region. Given the final box region, the local patch $I_{crop}$ is cropped from the image.

\textbf{Answering with Augmented Input.}
 The local patch $I_{crop}$ is enlarged by 1.5$\times$ and added into the image-question pair, forming a triples $(I, I_{crop}, q)$. Finally, the triples is fed into the LMM for auto-regressive generation. 
\section{Experiments}
\label{experiments}

\subsection{Evaluation Setup}
We report both the performance under CoT and direct answering settings. We select a variety of LMMs for evaluation, including both open-source and proprietary models. We also incorporate the models optimized for multimodal reasoning in the evaluation. 
\textbf{Proprietary Models:}
We select several cutting-edge proprietary models, including GPT-4o~\cite{gpt4o}, o4-mini~\cite{o4mini}, Claude-3.7-Sonnet~\cite{claude3.7sonnet}, and Gemini-2.5-Pro~\cite{gemini25pro}. 
\textbf{Open-Source Models:}
For open-source LMMs, we select a various of candidates including TextMonkey~\cite{liu2024textmonkey}, DocOwl2~\cite{mplug-docowl2}, DeepSeek-VL2~\cite{deepseekvl2}, NVILA~\cite{nvila}, Kimi-VL~\cite{kimivl}, Ovis2 series~\cite{ovis}, InternVL3 series~\cite{internvl3}, Qwen2.5-VL series~\cite{qwen2_5vl}, QvQ-72B-Preview~\cite{qvq}, and LLaVA-OV-1.5~\cite{llavaov1_5}.
For o4-mini, Claude-3.7-Sonnet, Gemini-2.5-Pro, and QvQ-72B-Preview, the max output token length is 8,192, while 2,048 is used for others. Considering the overall superior performance, LLaVA-OV-1.5 is selected as the baseline LMM for applying TextCue. Following ViCrop~\cite{mllm_know}, the fixed layer $m$ is 22. In adaptive layer selection (ALS), we set $n$ to 5.

\begin{figure*}[t!]
\centering
    \includegraphics[width=1\textwidth]{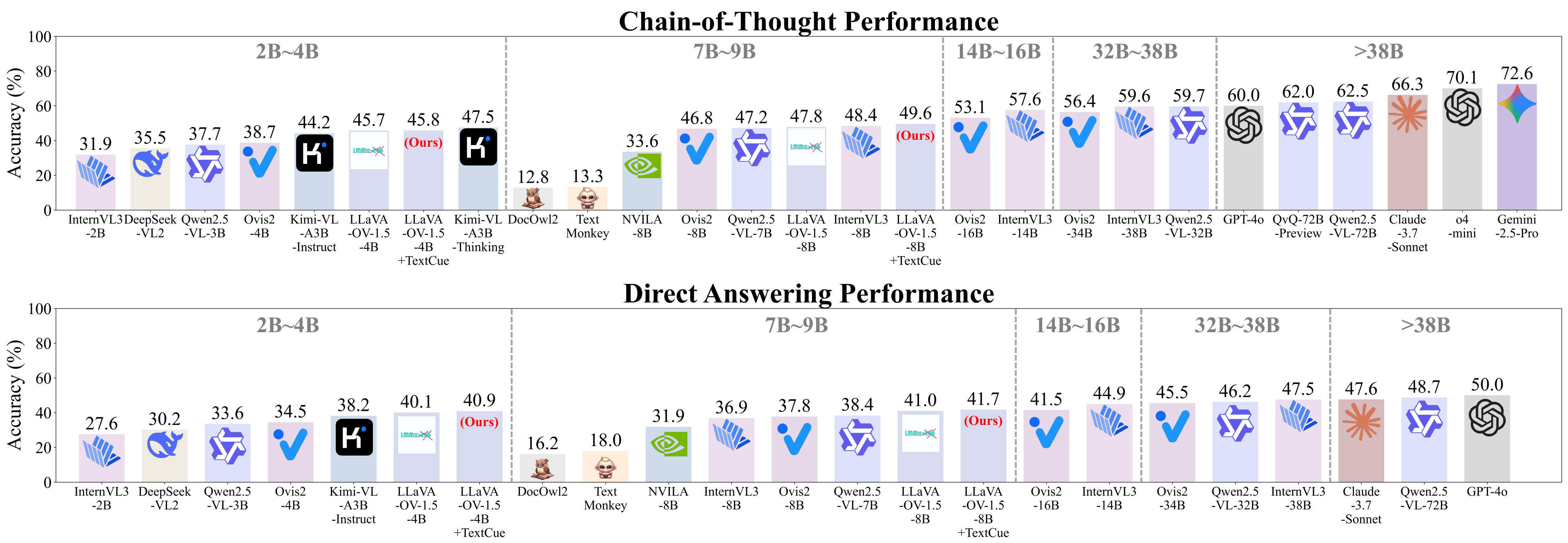}
    \caption{
    Comparison of LMMs on LogicOCR under CoT and direct answering settings. Applying TextCue on LLaVA-OV-1.5-4B achieves 0.1\% and 0.8\% accuracy improvements under CoT and direct answering settings, respectively, while achieving 1.8\% and 0.7\% accuracy gains on LLaVA-OV-1.5-8B under CoT and direct answering settings.
    }
    \label{fig:all_accuracy}
\end{figure*}

\subsection{Main Results}
The quantitative results of LMMs are displayed in Fig.~\ref{fig:all_accuracy} and Tab.~\ref{tab:3}. Overall, LogicOCR presents a significant challenge to most state-of-the-art LMMs. Under the CoT setting, Gemini-2.5-Pro achieves the highest accuracy of 72.4\% among both proprietary and open-source models. Recent open-source models have made notable strides in multimodal reasoning, with Qwen2.5-VL-72B outperforming GPT-4o. Applying TextCue on LLaVA-OV-1.5-4B achieves 0.1\% and 0.8\% accuracy gains under CoT and direct answering settings, respectively, while achieving 1.8\% and 0.7\% accuracy gains on LLaVA-OV-1.5-8B under CoT and direct answering settings.

\begin{table*}[t!]
\centering
\caption{
Detailed evaluation results on the subsets of LogicOCR under CoT and direct answering settings. In LogicOCR-Gen, questions may involve multiple reasoning types, with 17.4\%, 26.1\%, 35.0\%, and 21.5\% containing 1, 2, 3, and more than 3 reasoning types, respectively. `Num.', `Temp.', `De.', and `Cond.' represent numerical, temporal, decision, and conditional reasoning, respectively. `$\spadesuit$' denotes document-oriented LMMs.
}
\setlength{\tabcolsep}{2pt}
\resizebox{1\linewidth}{!}{\begin{tabular}{l|ccccc|c|ccccc|c}
\hline
\toprule
\multirow{2}{*}{\textbf{Model}} &\multicolumn{6}{c|}{\textbf{LogicOCR-Gen}} &\multicolumn{6}{c}{\textbf{LogicOCR-Real}} \\
\cline{2-7} \cline{8-13} 
&\multicolumn{4}{c}{Reasoning Types} &\textbf{Average} &\textbf{Average} &\multicolumn{4}{c}{Reasoning Types} &\textbf{Average} &\textbf{Average} \\
\cline{2-5} \cline{8-11} 
&1 &2 &3 &$>$3 &\textbf{(CoT)} &\textbf{(Direct)} &Num. &Temp. &De. &Cond. &\textbf{(CoT)} &\textbf{(Direct)} \\

\midrule

\rowcolor{gray!15} \multicolumn{13}{c}{\textit{Proprietary LMMs}} \\
GPT-4o~\cite{gpt4o} &64.6 &67.2 &60.5 &61.0 &63.1 &60.3 &54.7 &61.5 &57.7 &67.6 &58.0 &\textbf{43.2} \\
o4-mini~\cite{o4mini} &\underline{79.7} &\underline{77.0} &\underline{76.4} &\underline{77.1} &\underline{77.3} &$-$ &\underline{62.4} &\underline{68.4} &66.3 &\underline{73.8} &\underline{65.5} &$-$ \\
Claude-3.7-Sonnet~\cite{claude3.7sonnet} &75.0 &73.5 &68.1 &73.3 &71.8 &59.1 &58.8 &62.5 &\underline{69.3} &73.6 &62.6 &\underline{40.0} \\
Gemini-2.5-Pro~\cite{gemini25pro} &\textbf{82.3} &\textbf{81.7} &\textbf{80.6} &\textbf{78.4} &\textbf{80.7} &$-$ &\textbf{64.2} &\textbf{69.0} &\textbf{72.1} &\textbf{75.4} &\textbf{67.3} &$-$ \\

\rowcolor{gray!15} \multicolumn{13}{c}{\textit{Open-Source LMMs}} \\

TextMonkey~\cite{liu2024textmonkey} $\spadesuit$ &12.0 &15.3 &14.8 &22.5 &16.1 &24.0 &7.5 &10.9 &18.3 &23.0 &11.5 &14.2 \\
DocOwl2~\cite{mplug-docowl2} $\spadesuit$ &13.0 &22.6 &16.9 &22.9 &19.0 &16.2 &4.7 &6.3 &18.3 &20.6 &8.7 &16.2 \\
DeepSeek-VL2~\cite{deepseekvl2} &37.0 &36.6 &35.8 &39.8 &37.1 &41.3 &30.2 &29.3 &43.3 &48.3 &34.4 &23.0 \\
NVILA-8B~\cite{nvila} &37.0 &40.4 &39.0 &46.2 &40.6 &41.5 &23.4 &26.4 &36.5 &47.0 &29.0 &25.7 \\
Kimi-VL-A3B-Instruct~\cite{kimivl} &48.4 &49.8 &48.8 &51.3 &49.5 &54.1 &36.2 &35.6 &51.9 &54.8 &40.6 &27.8 \\
Kimi-VL-A3B-Thinking~\cite{kimivl} &55.7 &51.6 &53.8 &56.8 &54.2 &$-$ &39.2 &46.5 &44.2 &53.9 &43.1 &$-$ \\
\arrayrulecolor{gray!15}
\midrule
Ovis2-4B~\cite{ovis} &41.1 &41.8 &35.8 &41.1 &39.5 &45.0 &33.0 &32.2 &50.0 &54.8 &38.2 &27.5 \\
Ovis2-8B~\cite{ovis} &50.0 &50.5 &45.5 &47.0 &47.9 &49.2 &42.4 &43.7 &52.9 &57.3 &46.0 &30.4 \\
Ovis2-16B~\cite{ovis} &57.3 &57.5 &53.8 &50.4 &54.6 &52.6 &48.4 &47.7 &57.7 &65.1 &52.1 &34.3 \\
Ovis2-34B~\cite{ovis} &61.5 &56.1 &56.9 &55.9 &57.3 &59.8 &52.6 &52.3 &59.6 &67.6 &55.9 &36.1 \\
\midrule
InternVL3-2B~\cite{internvl3} &38.0 &37.6 &33.0 &32.2 &34.9 &37.7 &25.3 &23.0 &40.4 &46.1 &30.0 &20.9 \\
InternVL3-8B~\cite{internvl3} &58.9 &50.9 &48.1 &46.6 &50.4 &50.6 &44.6 &44.2 &53.8 &55.4 &47.2 &27.9 \\
InternVL3-14B~\cite{internvl3} &64.1 &61.3 &60.5 &58.5 &60.9 &61.6 &51.7 &54.6 &60.6 &67.0 &55.5 &33.9 \\
InternVL3-38B~\cite{internvl3} &64.6 &60.6 &59.7 &61.9 &61.3 &61.7 &55.2 &56.9 &60.6 &70.1 &58.6 &38.3 \\
\midrule
Qwen2.5-VL-3B~\cite{qwen2_5vl} &47.4 &47.4 &42.9 &43.2 &44.9 &49.5 &27.9 &29.9 &40.4 &49.5 &33.0 &23.1 \\
Qwen2.5-VL-7B~\cite{qwen2_5vl} &58.9 &50.2 &49.6 &49.6 &51.4 &52.6 &39.6 &44.2 &54.8 &57.6 &44.5 &29.2 \\
Qwen2.5-VL-32B~\cite{qwen2_5vl} &67.2 &63.4 &66.8 &63.1 &65.2 &\underline{64.2} &53.2 &49.4 &65.4 &66.7 &56.1 &34.5 \\
Qwen2.5-VL-72B~\cite{qwen2_5vl} &70.3 &64.8 &62.9 &67.4 &65.6 &\textbf{67.2} &57.4 &58.0 &61.5 &71.3 &60.4 &36.6 \\
QvQ-72B-Preview~\cite{qvq} &74.3 &71.0 &67.3 &70.2 &70.1 &$-$ &53.9 &54.6 &60.6 &65.9 &56.7 &$-$ \\
\midrule
LLaVA-OV-1.5-4B~\cite{llavaov1_5} &53.1 &50.2 &50.4 &52.5 &51.3 &58.4 &38.0 &41.9 &51.0 &53.0 &42.1 &28.0 \\
\rowcolor{gray!15} LLaVA-OV-1.5-4B+TextCue &50.5 &54.4 &49.6 &54.7 &52.1 &58.7 &37.0 &42.0 &46.2 &55.8 &41.7 &29.2 \\
LLaVA-OV-1.5-8B~\cite{llavaov1_5} &53.1 &52.6 &55.3 &53.0 &53.7 &59.4 &40.1 &40.2 &46.1 &58.2 &44.0 &28.9 \\
\rowcolor{gray!15} LLaVA-OV-1.5-8B+TextCue &59.9 &58.5 &54.3 &56.3 &56.8 &60.2 &40.7 &40.8 &52.9 &58.6 &44.9 &29.6\\

\arrayrulecolor{black}
\bottomrule
\hline
\end{tabular}}
\label{tab:3}
\vspace{-4mm}
\end{table*}

\textbf{Most LMMs Show No Improvement with CoT on LogicOCR-Gen}.
As shown in Tab.~\ref{tab:3}, we find that only large-scale LMMs, like Ovis2-16B, Qwen2.5-VL-32B, and proprietary models, benefit from CoT. In contrast, smaller models such as Ovis2-4B, InternVL3-2B, and Qwen2.5-VL-3B experience a notable performance drop when using CoT. This suggests that smaller LMMs lack the reliable multimodal reasoning capabilities needed for CoT, which may introduce unnecessary complexity, reducing their accuracy. For larger models, such as Ovis2-34B and InternVL3-38B, CoT still does not outperform direct answers, indicating that these models struggle to identify the appropriate reasoning path for logical tasks. Previous research~\cite{sprague2024cot} shows that CoT is more effective for math, symbolic, and algorithmic tasks, while its utility may be limited for tasks like commonsense reasoning. The LogicOCR-Gen subset presents a challenging multimodal reasoning scenario.

\textbf{Decision Reasoning Exhibits the Lowest Improvement while Model Scaling}. In Tab.~\ref{tab:3}, we observe that LMMs fall short on numerical, temporal, and decision reasoning. As model parameters scale up, the numerical and temporal reasoning performance are significantly enhanced. However, the decision reasoning exhibits the lowest improvement. For example, the numerical and temporal performance of Ovis2 are improved by 19.6\% and 20.1\% from 4B scale to 34B, while the decision reasoning performance is only enhanced by 9.6\%. It indicates that future LMMs should enhance their decision-making capabilities beyond robust reading and comprehension procedures, which is crucial for the agent application in real-world scenarios.

\begin{figure*}[t!]
\centering
    \includegraphics[width=1\textwidth]{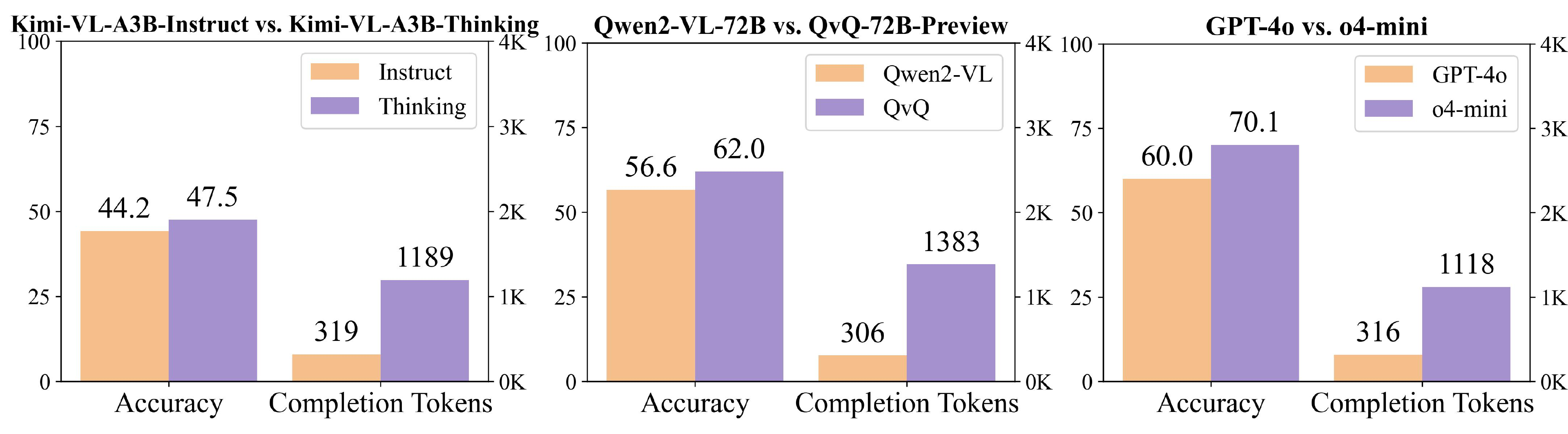}
    \caption{
    Comparison of average accuracy and output length (completion tokens) between general LMMs and their reasoning-enhanced counterparts. Notably, o4-mini achieves much higher accuracy using fewer tokens than QvQ-72B-Preview.
    }
    \label{fig:test_time_scaling}
\end{figure*}

\subsection{Analysis}

\subsubsection{Is Test-Time Scaling Beneficial for LogicOCR?}
Recent models have explored test-time scaling to enhance multimodal reasoning. Our experiments reveal that \textbf{test-time scaling significantly improves performance on LogicOCR, though the efficiency of open-source LMMs still leaves room for improvement}. 
Specifically, for open-source models, we select Kimi-VL-A3B-Thinking~\cite{kimivl} and QvQ-72B-Preview~\cite{qvq}, and compare them with their based counterparts, \ie, Kimi-VL-A3B-Instruct and Qwen2.5-VL-72B. 
For proprietary models, we assess o4-mini~\cite{o4mini}. Figure~\ref{fig:test_time_scaling} presents the average accuracy under the CoT setting and the corresponding completion token usage.
Kimi-VL-A3B-Thinking achieves a 3.3-point accuracy gain over its base version, Kimi-VL-A3B-Instruct, but requires 3.7$\times$ more completion tokens. It is noteworthy that o4-mini attains 8.1 points higher accuracy than QvQ-72B-Preview while using only 80\% of its token length, suggesting potential redundancy in QvQ-72B-Preview's reasoning.
These observations indicate that future reasoning LMMs should aim for more concise reasoning strategies to enhance efficiency without compromising accuracy. 
In terms of data and training, curating variable-length CoT data according to task difficulty is essential~\cite{seed1.5-vl}. For reinforcement learning, applying length penalty on responses while ensuring the answer correctness is a natural and useful solution. 

\begin{figure*}[t!]
\centering
    \includegraphics[width=1\textwidth]{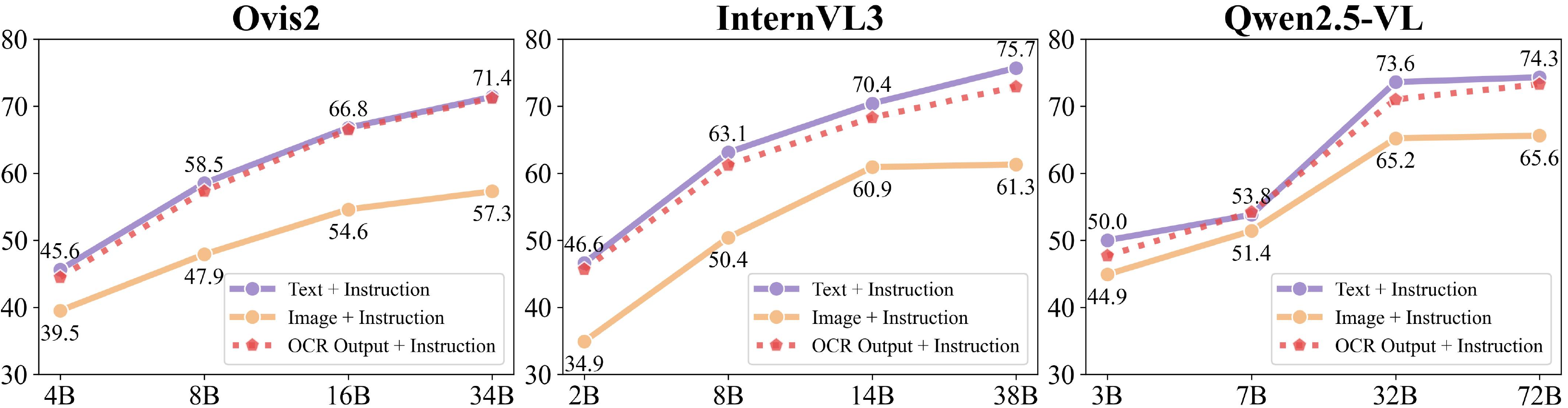}
    \vspace{-4mm}
    \caption{
    Impact of input modalities on LMMs with CoT on LogicOCR-Gen. These LMMs use Qwen2.5~\cite{qwen2.5} as the backbone. `Text + Instruction' denotes text-only input, where the question is provided in the text. `Image + Instruction' refers to multimodal input, with the question embedded in the image. `OCR Output + Instruction' represents feeding LMM's OCR-extracted text to themselves instead of ground-truth text. 
    While this two-step strategy (prompt given in Appendix D) yields higher accuracy than direct multimodal input, it incurs significant inference overhead and contradicts the objective of end-to-end multimodal reasoning from raw visual inputs without task-specific priors.
    }
    \label{fig:modal_curve}
    \vspace{-2mm}
\end{figure*}

\begin{table*}[t!]
\centering
\caption{
OCR performance of LMMs on LogicOCR. The used prompt is shown in Appendix C. 
The collected text corpus serves as ground truth for OCR. Results show strong OCR capabilities across models. \textit{The strong performance also highlights the high visual-text fidelity of GPT-Image-1}.
}
\setlength{\tabcolsep}{8pt}
\resizebox{1\linewidth}{!}{\begin{tabular}{l|cccccc}
\hline
\toprule
\textbf{Model} &\textbf{Edit Distance$\downarrow$} &\textbf{F1-score$\uparrow$} &\textbf{Precision$\uparrow$} &\textbf{Recall$\uparrow$} &\textbf{BLUE$\uparrow$} &\textbf{METEOR$\uparrow$} \\
\midrule
\arrayrulecolor{gray!80}
Ovis2-4B~\cite{ovis} &0.052 &94.8 &95.3 &94.5 &90.0 &94.1 \\
Ovis2-8B~\cite{ovis} &0.023 &97.4 &98.1 &96.8 &94.1 &96.6 \\
Ovis2-16B~\cite{ovis} &\underline{0.020} &\underline{97.6} &\underline{98.2} &\underline{97.1} &\underline{94.5} &\underline{96.9} \\
Ovis2-34B~\cite{ovis} &\textbf{0.020} &\textbf{97.8} &\textbf{98.3} &\textbf{97.3} &\textbf{94.9} &\textbf{97.1} \\

\midrule
InternVL3-2B~\cite{internvl3} &0.048 &95.6 &97.2 &94.6 &90.4 &94.3 \\
InternVL3-8B~\cite{internvl3} &0.026 &96.8 &97.4 &96.4 &92.9 &96.2 \\
InternVL3-14B~\cite{internvl3} &\underline{0.021} &\underline{96.9} &\underline{97.5} &\underline{96.5} &\underline{93.3} &\underline{96.3} \\
InternVL3-38B~\cite{internvl3} &\textbf{0.020} &\textbf{97.1} &\textbf{97.6} &\textbf{96.6} &\textbf{93.6} &\textbf{96.5} \\

\midrule
Qwen2.5-VL-3B~\cite{qwen2_5vl} &0.041 &96.5 &\textbf{98.1} &95.5 &92.1 &94.9 \\
Qwen2.5-VL-7B~\cite{qwen2_5vl} &\underline{0.032} &\underline{97.0} &\underline{98.0} &96.2 &92.9 &95.6 \\
Qwen2.5-VL-32B~\cite{qwen2_5vl} &0.034 &96.8 &96.6 &\textbf{97.0} &\textbf{93.4} &\textbf{96.6} \\
Qwen2.5-VL-72B~\cite{qwen2_5vl} &\textbf{0.028} &\textbf{97.1} &97.8 &\underline{96.5} &\underline{93.2} &\underline{96.0} \\

\arrayrulecolor{black}
\bottomrule
\hline
\end{tabular}}
\label{tab:4}
\vspace{-4mm}
\end{table*}

\subsubsection{Have LMMs Effectively Bridged Visual Reading and Reasoning?} 
To investigate this question, we evaluate Ovis2~\cite{ovis}, InternVL3~\cite{internvl3}, and Qwen2.5-VL~\cite{qwen2_5vl} series on LogicOCR-Gen. Each model is tested with text-only and multimodal inputs to assess whether their reasoning over visual content matches their language-only reasoning performance. Additionally, we test LMMs in text-only setting with their OCR results as input, instead of ground-truth text.

As shown in Fig.~\ref{fig:modal_curve}, all models exhibit a noticeable drop in reasoning performance when processing multimodal inputs. For instance, the Ovis2 series achieves, on average, 10.8 points higher accuracy with text-only input. InternVL3 shows a larger average gap of 12.1 across scales. In contrast, Qwen2.5-VL demonstrates better alignment between modalities, with an average gap of just 6.2, and only 2.4 for the Qwen2.5-VL-7B variant. This stronger modal consistency likely stems from more extensive vision-language pre-training~\cite{qwen2_5vl}. Notably, the performance gap does not appear to result from deficiencies in visual recognition, as state-of-the-art models perform well in the OCR task (see Tab.~\ref{tab:4}). Moreover, in the text-only setting, using either ground-truth question context or OCR predictions from LMMs yields similar accuracy (see purple and red dashed lines in Fig.~\ref{fig:modal_curve}), suggesting that current LMMs have yet to fully integrate visual reading with high-level reasoning.

Moreover, a closer examination of model scaling reveals that both input modalities, text and multimodal, benefit from increased model capacity, primarily due to enhanced reasoning capabilities. For example, as model size increases from 8B to 38B parameters, InternVL3 improves multi-choice accuracy by 12.6 points with text input and 10.9 points with multimodal input, while OCR performance remains largely unchanged (Tab.~\ref{tab:4}). However, when scaling from \textasciitilde14B to \textasciitilde30B parameters for Ovis2 and InternVL3, accuracy gains under multimodal input diminish, whereas improvements with text-only input remain notable. This suggests that aligning vision and language representations for multimodal reasoning remains a significant challenge at larger scales.

These findings highlight two key insights:
First, \textbf{LMMs still fall short of fully integrating visual reading and reasoning}.
Second, while vision-language alignment suffices for perception tasks like OCR, \textbf{it remains inadequate for more complex reasoning, especially as model size grows}. 
Therefore, achieving thorough vision-language alignment is vital for advancing multimodal reasoning in the future. While TokenVL~\cite{guan2025token} explore explicit alignment between visual and language tokens through contrastive learning and text mask segmentation, there remains an opportunity to design scalable and tailored training objectives and datasets for reasoning tasks. In addition, perhaps we can explore an alignment feedback (\eg, performance gap between different modalities) to boost the self-improvement on vision-language integration.

\subsubsection{Towards Robust LMMs Against Visual-Text Orientation Variations}
\begin{figure*}[t!]
\centering
    \includegraphics[width=1\textwidth]{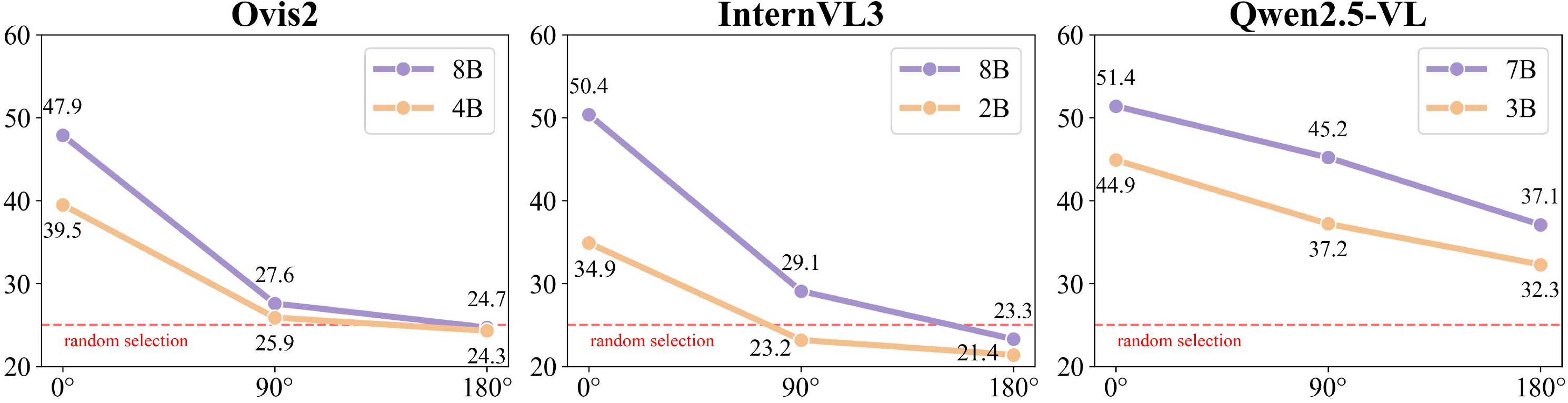}
    \caption{
    Impact of visual-text orientation on LogicOCR-Gen. Images were rotated 90$\degree$ and 180$\degree$ clockwise to alter text orientation. Results show that state-of-the-art LMMs are sensitive to such changes, \eg, Ovis2 and InternVL2 accuracy drops to near-random levels, while Qwen2.5-VL demonstrates greater robustness.
    }
    \label{fig:rotation_curve}
\end{figure*}

We observe that recent state-of-the-art LMMs exhibit sensitivity to visual-text orientation. As illustrated in Fig.~\ref{fig:rotation_curve}, rotating input images sharply degrades performance, \eg, Ovis2-4B and InternVL3-2B accuracy drops by 13.6 and 11.7 points, respectively, when rotated 90$\degree$. Larger models fare worse: Ovis2-8B and InternVL3-8B show declines of 20.3 and 21.3 points. In contrast, the Qwen2.5-VL series demonstrates greater robustness, maintaining stable performance even with 180$\degree$ rotations.

In summary, \textbf{the perception robustness of LMMs across different visual-text orientations needs further improvement}. Qwen2.5-VL outperforms others in this evaluation, likely due to data augmentation strategies during training, such as image rotation, which is commonly employed to enhance the robustness of OCR and visual-text parsing model. 
For better robustness on varied visual-text orientations, introducing rotation-equivariant image encoder may be helpful.

\subsubsection{Error Analysis}
On LogicOCR-Gen, we analyze the error types of Qwen2.5-VL-72B during reasoning by categorizing them into six major types: 1) conceptual error, 2) logistic error, 3) argument structure error, 4) option analysis error, 5) information usage error, and 6) image reading error. These categories encompass 17 specific error types in total, though not all occur in every failure case. Definitions of each error type can be found in Appendix E. On LogicOCR-Real, the major error types are: 1) logistic error, 2) computation error, 3) information usage error, and 4) image reading error. To aid in analysis, we use o4-mini~\cite{o4mini}, with the prompt template provided in Appendix E. Our analysis is based on 191 instances from LogicOCR-Gen and 194 instances from LogicOCR-Real where Qwen2.5-VL-72B fails, but o4-mini answers correctly.

\begin{figure*}[t!]
\centering
  \includegraphics[width=1\textwidth]{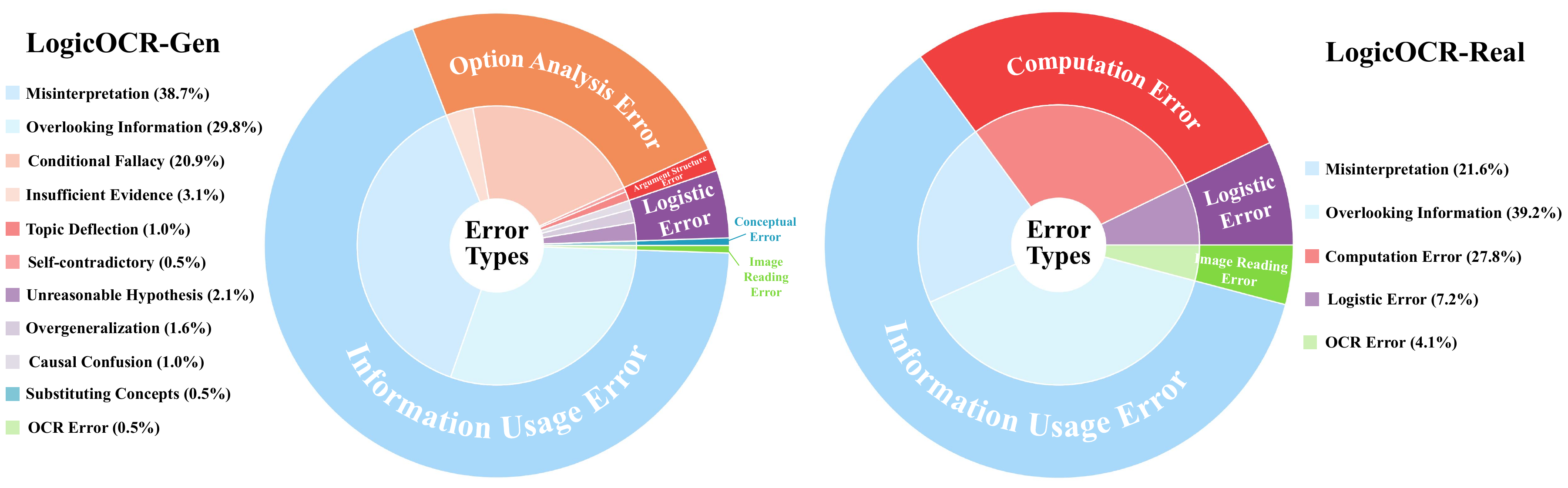}
  \caption{
  Error analysis for Qwen2.5-VL-72B reveals four main issues: misinterpretation, overlooking key information, computation error, and conditional fallacy.
  }
  \vspace{-2mm}
  \label{fig:error_analysis}
\end{figure*}

As shown in Fig.~\ref{fig:error_analysis}, the main errors in the reasoning process are information usage, option analysis, and computation errors.
First, information usage errors occur when the model misinterprets the context, leading to an incorrect reasoning path. It often overlooks key information or requirements in multimodal context.
Second, option analysis errors on LogicOCR-Gen are mainly due to conditional fallacy, where the model struggles to distinguish between necessary and sufficient conditions or gets confused by affirmative and negative forms. Additionally, some choices are selected without sufficient evidence.
Finally, computation errors on LogicOCR-Real occur when the questions involve numerical calculation. Numerical calculation is still a crucial issue for mulimodal reasoning.
In the future, several strategies can be explored to mitigate these errors and improve overall reasoning performance. For example, integrating retrieval-augmented models may help ensure that all relevant details are retained and revisited throughout the reasoning process. In addition, during training, we can categorize training instances by error type via annotations or model feedback, then dynamically resample training data to improve under-performing categories.

\subsection{Ablation Studies on TextCue}

\begin{wraptable}[6]{r}{7cm}
\centering
\setlength{\tabcolsep}{8pt}
\vspace{-8mm}
\caption{
Ablation results of TextCue.
}
\resizebox{7cm}{!}{
\begin{tabular}{l|cc}
\hline
\toprule
\textbf{Method} &\textbf{CoT} &\textbf{Direct} \\
\midrule
LLaVA-OV-1.5-8B &47.8 &41.0 \\
LLaVA-OV-1.5-8B+ViCrop &49.0 &41.0 \\
LLaVA-OV-1.5-8B+ViCrop+ALS &\underline{49.2} &\underline{41.3} \\
LLaVA-OV-1.5-8B+TextCue &\textbf{49.6} &\textbf{41.7} \\

\bottomrule
\hline

\end{tabular} }
\label{tab:ablation}
\end{wraptable}

\textbf{Influence of the modifications over baseline training-free method}.
As shown in Tab.~\ref{tab:ablation}, we investigate the influence of adaptive layer selection (ALS) scheme and integrating the layout predictions of text segmentation model~\cite{hisam} over the baseline training-free method, ViCrop~\cite{mllm_know}. Specifically, introducing ALS scheme improves 0.2\% and 0.3\% accuracy on CoT and direct-answering settings compared to ViCrop. Integrating the layout predictions further achieves 0.4\% accuracy improvements on both CoT and direct-answering performance. Overall, based on LLaVA-OV-1.5-8B, applying TextCue enhances the CoT and direct-answering accuracy by 1.8\% and 0.7\%, respectively.

\begin{figure*}[h]
\centering
    \includegraphics[width=1\textwidth]{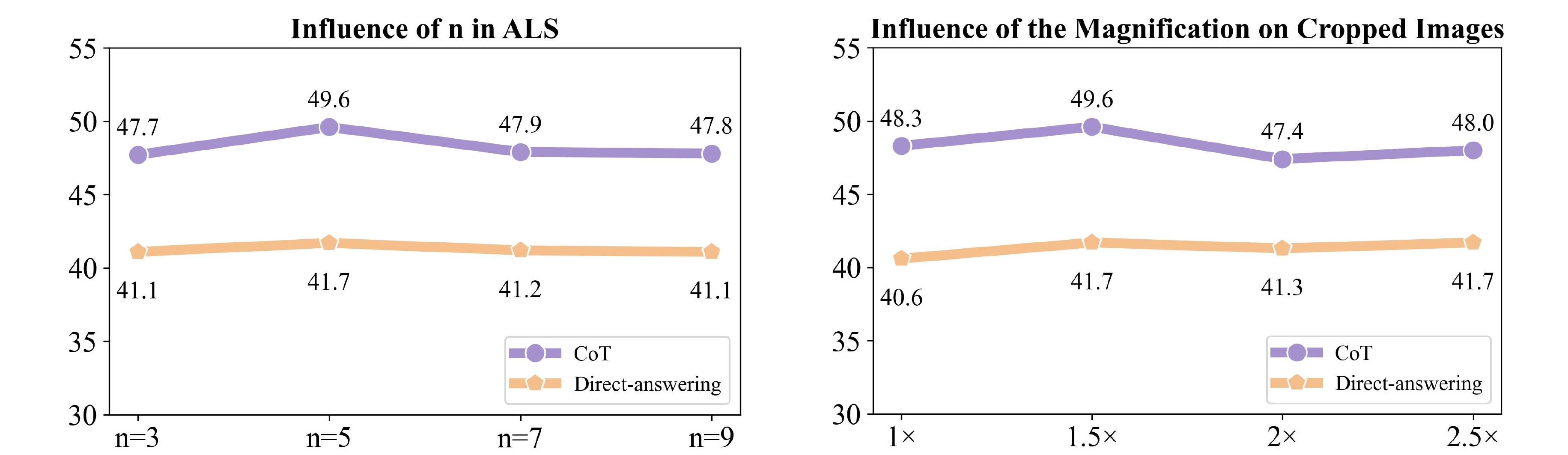}
    \caption{
    Influence of the number ($n$) of attention maps involved in ALS (left) and the magnification on cropped images (right). LLaVA-OV-1.5-8B is used.
    }
    \label{fig:ablation}
\end{figure*}

\textbf{Influence of the number of attention maps involved in ALS scheme}. 
As shown in Fig.~\ref{fig:ablation}, selecting $n=5$ achieves the best accuracy under both CoT and direct answering settings on LogicOCR.  

\textbf{Influence of the cropped image magnification}. 
We investigate the influence of the cropped image magnification in Fig.~\ref{fig:ablation}. Overall, enlarging the cropped images by 1.5$\times$ delivers the best accuracy under both CoT and direct answering settings. While upscaling cropped image regions by higher factors of 2$\times$ and 2.5$\times$, the expansion process introduces an increased density of local patch features, which may distort the contextual information for LMM, leading to the performance decline and additional computation overhead.

\subsection{Applicability of TextCue on Other LMMs}

\begin{wraptable}[14]{r}{6cm}
\centering
\setlength{\tabcolsep}{8pt}
\vspace{-8mm}
\caption{
TextCue results with Qwen2.5-VL and Ovis2.5.
}
\resizebox{6cm}{!}{
\begin{tabular}{l|cc}
\hline
\toprule
\textbf{Method} &\textbf{CoT} &\textbf{Direct} \\
\midrule
Qwen2.5-VL-3B~\cite{qwen2_5vl} &\textbf{37.7} &33.6 \\
Qwen2.5-VL-3B+ViCrop &35.6 &\underline{34.0} \\
Qwen2.5-VL-3B+TextCue &\underline{36.5} &\textbf{34.5} \\
\midrule
Qwen2.5-VL-7B~\cite{qwen2_5vl} &\textbf{47.2} &38.4 \\
Qwen2.5-VL-7B+ViCrop &44.6 &\underline{38.5} \\
Qwen2.5-VL-7B+TextCue &\underline{45.0} &\textbf{38.6} \\
\midrule
Ovis2.5-2B~\cite{lu2025ovis2_5} &42.1 &33.8 \\
Ovis2.5-2B+ViCrop &\underline{42.7} &\underline{34.1} \\
Ovis2.5-2B+TextCue &\textbf{42.8} &\textbf{34.3} \\
\midrule
Ovis2.5-9B~\cite{lu2025ovis2_5} &56.8 &43.5 \\
Ovis2.5-9B+ViCrop &\underline{57.0} &\underline{43.7} \\
Ovis2.5-9B+TextCue &\textbf{57.5} &\textbf{44.1} \\
\bottomrule
\hline

\end{tabular} }
\label{tab:ovis_qwen_textcue_res}
\end{wraptable}

We further apply TextCue and ViCrop~\cite{mllm_know} on Qwen2.5-VL-3B, Qwen2.5-VL-7B, Ovis2.5-2B, and Ovis2.5-9B. As shown in Tab.~\ref{tab:ovis_qwen_textcue_res}, our TextCue method achieves the best accuracy performance under direct answering setting and CoT setting with Ovis2.5 as the base LMM. For example, applying TextCue on Ovis2.5-9B achieves 0.7\% and 0.6\% higher accuracy under CoT and direct answering settings, respectively. 
However, we observe that applying visual cropping augmentation on Qwen2.5-VL under CoT answering leads to accuracy decline, which may stem from the inherent differences among LMMs. Visual cropping generates extra image tokens in addition to the original image, which could introduce unexpected information and disturb the attention distribution during the CoT reasoning process of Qwen2.5-VL. Nonetheless, our method obtains better performance than ViCrop on Qwen2.5-VL. 
Future works could explore leveraging multiple cues to strengthen the CoT answering accuracy and delivering better applicability on different LMMs.
\section{Limitations}
\label{Limitations}

First, the LogicOCR-Gen subset currently focuses on multi-choice questions. Although LogicOCR-Gen has revealed some useful insights, the data generation pipeline can be extended for free-form reasoning tasks while further efforts are required to collect the target corpora. 
Second, as for the visual cropping method, the overall performance remains bounded by the base LMM. As future LMMs continue to evolve with improved perception, reasoning, and visual agent capabilities, the improvement achieved by external training-free visual cropping strategy may gradually diminish. 
Finally, probably due to the variance among LMMs, the applicability of TextCue could be constrained to specific LMMs.

\section{Conclusion}
\label{Conclusion}

In this work, we present LogicOCR benchmark, consisting of LogicOCR-Gen and LogicOCR-Real subsets designed to evaluate the logical reasoning abilities of Large Multimodal Models (LMMs) on text-rich images, with minimal reliance on STEM knowledge. We develop a scalable pipeline that leverages GPT-Image-1 to generate visually diverse, contextually grounded images from a curated text corpus, followed by manual verification to ensure quality.
Our evaluation of comprehensive LMMs under both direct answering and Chain-of-Thought (CoT) settings reveals that most models do not benefit from CoT on LogicOCR-Gen, suggesting potential weaknesses in their reasoning processes. While these models excel at OCR, their performance on multimodal reasoning lags behind their text-only counterparts, indicating a gap between visual understanding and logical reasoning. We also show that LMMs are sensitive to visual-text orientation and benefit from test-time scaling, highlighting important factors affecting multimodal reasoning. Additionally, we offer a training-free method, TextCue, for improving the performance of baseline LMM.
We hope LogicOCR serves as a valuable resource for advancing research in OCR-related multimodal reasoning.

\section*{Acknowledgements}
This work was supported in part by the National Key Research and Development Program of China under Grant 2023YFC2705700, in part by the National Natural Science Foundation of China under Grants U23B2048 and 623B2076, in part by the Innovative Research Group Project of Hubei Province under Grant 2024AFA017, and in part by the Science and Technology Major Project of Hubei Province under Grants 2024BAB046 and 2025BCB026. The numerical calculations in this paper have been done on the supercomputing system in the Supercomputing Center of Wuhan University. This work was also supported by WHU-Kingsoft Joint Lab.

\appendix
\section{Prompt Templates for Text-to-Image Generation with GPT-Image-1}
\label{apdx:prompt4gptimage1}
The four prompt templates for instructing GPT-Image-1 are provided here. Note that \{context\}, \{question\}, and \{options\} are place-holders for the context, question, and options parts of each sample, respectively. And \{depiction\} is a place-holder for the background depiction generated by Qwen2.5.

\begin{promptbox}[Interleaved Style Layout \& Non-Handwritten Style Font]{violet}
Generate an image about random color paper with smallest font-size. Firstly, it is written about context information: "\{context\}"

An illustration figure or scene described by the above context is shown.

Then, the image displays a question: "\{question\}"

Finally, four choices are written on the image: "\{options\}"

Do not summarize the visual text content given above.
\end{promptbox}

\begin{promptbox}[Interleaved Style Layout \& Handwritten Style Font]{violet}
Generate an image about random color paper with smallest font-size. Firstly, it is written about context information in handwritten style: "\{context\}"

An illustration figure or scene described by the above context is shown.

Then, the image displays a question in handwritten style: "\{question\}"

Finally, four choices in handwritten style are written on the image: "\{options\}"

Do not summarize the visual text content given above.
\end{promptbox}

\begin{promptbox}[Background Style Layout \& Non-Handwritten Style Font]{violet}
Generate an image with smallest font-size. \{depiction\}
Some text paragraphs with contrastive color are shown. Specifically, firstly, it is written about context information: "\{context\}"

Then, the image displays a question: "\{question\}"

Finally, four choices are written on the image: "\{options\}"

Do not summarize the visual text content given above.
\end{promptbox}

\begin{promptbox}[Background Style Layout \& Handwritten Style Font]{violet}
Generate an image with smallest font-size. \{depiction\}
Some text paragraphs  with handwritten style and contrastive color are shown. Specifically, firstly, it is written about context information: "\{context\}"

Then, the image displays a question: "\{question\}"

Finally, four choices are written on the image: "\{options\}"

Do not summarize the visual text content given above.
\end{promptbox}

\section{Prompt Templates for Evaluating LMMs}
\label{apdx:prompt4eval}

The prompt templates used for LogicOCR-Gen under different settings are listed below.

\begin{promptbox}[Prompt for Multimodal Input under CoT setting]{myblue}
Solve the multi-choice question in image and then answer with one option letter. The last line of your response should be of the following format: `Answer: LETTER' where LETTER is one of options. Think step by step before answering.
\end{promptbox}

\begin{promptbox}[Prompt for Multimodal Input under Direct Answering Setting]{myblue}
Solve the multi-choice question in image. Directly answer the question with one option letter without explanation.
\end{promptbox}

\begin{promptbox}[Prompt for Pure Text Input under CoT setting]{myblue}
Question: \{context\} \{question\}

Options:

\{options\}

Solve the multi-choice question and then answer with one option letter. The last line of your response should be of the following format: 'Answer: LETTER' where LETTER is one of options. Think step by step before answering.
\end{promptbox}

\begin{promptbox}[Prompt for Pure Text Input under Direct Answering Setting]{myblue}
Question: \{context\} \{question\}

Options:

\{options\}

Directly answer the question with one option letter without explanation.
\end{promptbox}

The prompt templates used for LogicOCR-Real under different settings are listed below.

\begin{promptbox}[Prompt for Multimodal Input under CoT setting]{myblue}
\{question\} The last line of your response should be of the following format: 'Answer: YOUR\_ANSWER' where YOUR\_ANSWER is the final answer. Think step by step before answering.
\end{promptbox}

\begin{promptbox}[Prompt for Multimodal Input under Direct Answering Setting]{myblue}
\{question\} Directly answer the question using a single word or phrase without explanation.
\end{promptbox}

\section{Prompt Template for Evaluating OCR Performance on LogicOCR}
\label{apdx:ocr prompt}
\begin{promptbox}[The OCR Prompt for LMMs]{lightblue}
Please recognize the text paragraphs in the image. Do not add explanation. Do not answer the question in this image.
\end{promptbox}

\section{Prompt Template for Evaluating LMMs with Their OCR Results as Input}
\label{apdx:ocr then answer prompt}
In this evaluation, the image is not provided to LMMs. The OCR results are achieved in advance by using the prompt in Appendix~\ref{apdx:ocr prompt} and then inserted in the place-holder `\{OCR results\}', resulting in a two-step solution. Note that state-of-the-art LMMs have been equipped with strong OCR performance in order to eliminate the need for an explicit OCR step. However, we observe that end-to-end multimodal reasoning underperforms this time-consuming two-step strategy on LogicOCR, suggesting that state-of-the-art LMMs have yet to fully bridge visual reading with reasoning. This two-step strategy is actually impractical for real-world applications due to its substantial inference latency and the restricted assumption that an image can be represented by sole OCR results. Through this evaluation, we want to reveal one of the key bottlenecks in multimodal reasoning, rather than encouraging this two-step strategy.

\begin{promptbox}[Prompt for Pure Text Input under CoT Setting.]{myblue}
\{OCR results\}

\vspace{3mm}
Solve the above multi-choice question and then answer with one option letter. The last line of your response should be of the following format: 'Answer: LETTER' where LETTER is one of options. Think step by step before answering.
\end{promptbox}

\section{The Prompt Template for Error Analysis with o4-mini}
\label{apdx:error_analysis}

In the prompt, \{corpus\}, \{solution\}, and \{response\} are place-holders for the text corpus of each question, the correct option's letter, and the response of evaluated model.

\begin{promptbox}[Prompt for Error Analysis on LogicOCR-Gen]{orrange}
Here is a multi-choice question written on image:

\{corpus\}

\vspace{3mm}
The correct choice is \{solution\}. After reading the image, one AI model got a wrong answer process:

\{response\}

\vspace{3mm}
Please carefully analyze why the process is wrong and choose one most appropriate error from the following 17 types. Answer the error type of the following format "error type: chosen-error", such as "error type: 6. overgeneralization".

\vspace{3mm}
\#\# about conceptual error

1. substituting concepts: Failure to maintain conceptual consistency within the same reasoning.

2. improper juxtaposition: Confusing classification standards.

3. circular definition: The defining term directly or indirectly includes the defined term, such as "an optimist is an optimistic person."

\#\# about logistic error

4. unreasonable hypothesis: Adding subjective assumptions or reverse reasoning (e.g., from "smokers are in poor health" to "non-smokers are in good health").

5. exaggeration: Expressing possibility as certainty, such as exaggerating "may be extinct" to "already extinct".

6. overgeneralization: Using local phenomena to infer the whole, such as using "14 percent people like Peking Opera" to infer "general lack of traditional culture".

7. causal confusion: Reversing or imposing causal relationships, such as mistaking "low immune system causes psychological problems" for "psychological problems lower immunity."

\#\# about argument structure error

8. topic deflection: Deviating from the original discussion focus.

9. self-contradictory: Affirming contradictory propositions at the same time, such as "entangled at all times" and "temporarily put aside".

10. equivocating: There is no clear statement on right and wrong issues, such as "neither comprehensive nor one-sided".

11. circular argument: The argument relies on the premise itself, such as "lying is treasonous, therefore you are a traitor."

\#\# about option analysis error

12. overstatement: It exceeds the reasonable scope of the question, such as inferring "may" as "certain".

13. conditional fallacy: Can't distinguish between necessary and sufficient conditions, or confuse affirmative and negative forms.

14. insufficient evidence: The options lack support from the question or the information is one-sided.

\#\# about information usage error

15. misinterpretation: The key words in the question (such as "most different") are not captured.

16. overlooking information: Omission of key information in the material leads to misjudgment.

\#\# about image reading error

17. OCR error: The wrong optical character recognition results from the image affect the reasoning process.
\end{promptbox}

\begin{promptbox}[Prompt for Error Analysis on LogicOCR-Real]{orrange}
Here is a question based on the image:

\{question\}

\vspace{3mm}
The correct answer is \{solution\}. After reading the image, one AI model got a wrong answer process:
\{response\}

\vspace{3mm}
Please carefully analyze why the process is wrong and choose one most appropriate error from the following 17 types. Answer the error type of the following format "error type: chosen-error", such as "error type: 1. logical error".

\vspace{3mm}
\#\# about logic error

1. logical error: The model may not have performed the inference steps correctly, or may have produced inconsistent or unreasonable conclusions during inference.

\#\# about numerical error

2. computation error: The question involves numerical calculation and the model produces errors during calculation.

\#\# about information usage error

3. misinterpretation: The key words in the question (such as "most different") are not captured.

4. overlooking information: Omission of key cues in the image leads to misjudgment.

\#\# about image reading error

5. OCR error: The wrong optical character recognition results from the image affect the reasoning process.

\end{promptbox}

\section{Prompt Reference for Determining Layout Style}
\begin{promptbox}[Prompt for Choosing Layout Style]{myred}
Imagine you are designing an image. Given the context information between <context> and </context> tags. Is it more suitable to draw a full frame background scene or an illustration figure? Answer "background" or "illustration".

<context>\{context\}</context>
\end{promptbox}

\section{Prompt Reference for Background Depiction}
\begin{promptbox}[Prompt for Background Depiction]{myred}
You are designing the background scene for an image based on the context information given between <context> and </context> tags, please generate a short paragraph of caption in pure English for the background scene.

<context>\{context\}</context>
\end{promptbox}

\section{Comparison of Relative Attentions}
\label{sec:rel_attn_comp}

\begin{figure*}[!h]
    \centering
    \includegraphics[width=1\linewidth]{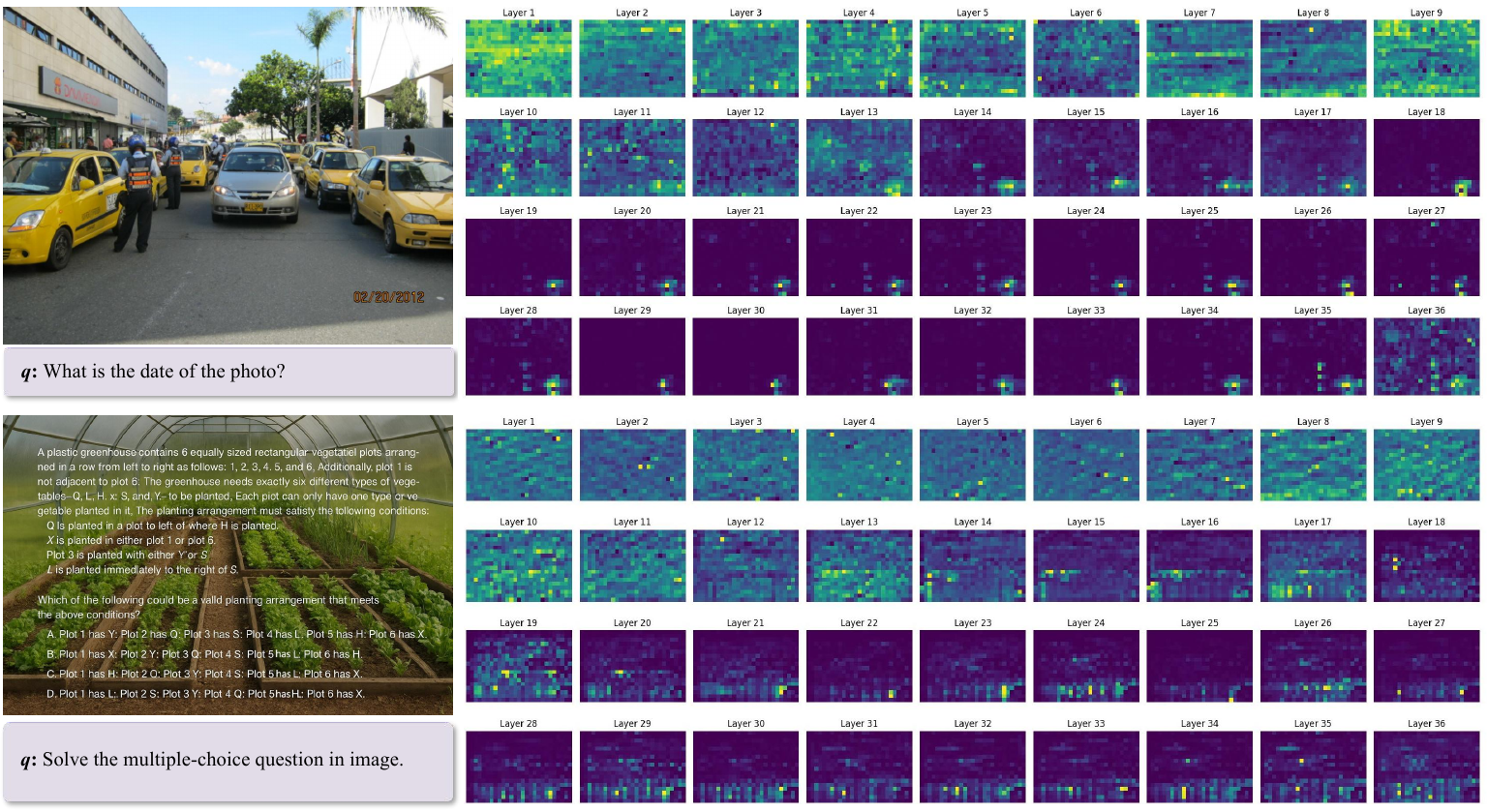}
    \caption{
    Comparison of relative attentions between a simple question (top) and a complex reasoning question (bottom). After the middle layer of LMM (LLaVA-OV-1.5 8B is used here), the attentions highlight the region where the answering cue is located. The attention maps for simple question are focused and exhibit higher similarity between each other, whereas the attention maps for reasoning problem on the text-rich image show the opposite pattern.
    }
    \label{fig:rel_attn_comp}
\end{figure*}

As shown in Fig.~\ref{fig:rel_attn_comp}, after the middle layer, the relative attentions highlight the region where the answering cue is located. The attention maps for simple question are focused and exhibit higher similarity between each other, whereas the attention maps for reasoning problem show the opposite pattern. 
Hence, we introduce the adaptive layer selection scheme to choose the most salient relative attention map.

\section{Broader Impacts}

This work contributes the LogicOCR benchmark that highlights critical limitations in current LMMs with respect to complex logical reasoning over text-rich images. Our evaluation of various LMMs under both direct-answer and CoT settings reveals that most models do not benefit from CoT prompting, suggesting potential shortcomings in their reasoning processes. While state-of-the-art LMMs excel at OCR, their performance on multimodal reasoning lags behind their text-only counterparts, indicating a gap between visual reading and logical reasoning. Additionally, our analysis shows that LMMs' performance is sensitive to visual-text orientation. Identifying these key flaws is crucial, especially in document understanding tasks and high-stakes applications such as medical and healthcare domains, where erroneous reasoning can lead to significant consequences. By revealing these limitations, LogicOCR provides a valuable source for guiding future improvements in multimodal reasoning. As an evaluation benchmark, there is no direct negative societal impacts.

\section{Datasheet for LogicOCR}

\subsection{Motivation}

\noindent \textbf{1. For what purpose was the dataset created? Was there a specific task in mind? Was there a specific gap that needed to be filled? Please provide a description.}

\textbf{A1:}  Existing multimodal reasoning datasets often require extensive mathematical or scientific knowledge, making it difficult to isolate pure reasoning ability from domain expertise. In contrast, most OCR-related benchmarks lack complexity and some of them only focus on narrow topics. LogicOCR was created to evaluate LMMs' complex logical reasoning capabilities on text-rich images while minimizing reliance on STEM knowledge.

\noindent \textbf{2. Who created this dataset (e.g., which team, research group) and on behalf of which entity (e.g., company, institution, organization)?}

\textbf{A2:} This dataset is created by the authors of this paper.

\noindent \textbf{3. Who funded the creation of the dataset? If there is an associated grant, please provide the name of the grantor and the grant name and number.}

\textbf{A3:} N/A.

\subsection{Composition}

\noindent \textbf{1. What do the instances that comprise the dataset represent (e.g., documents, photos, people, countries)? Are there multiple types of instances(e.g., movies, users, and ratings; people and interactions between them; nodes and edges)? Please provide a description.}

\textbf{A1:} Each multimodal sample in LogicOCR consists of one image and a question.

\noindent \textbf{2. How many instances are there in total (of each type, if appropriate)?}

\textbf{A2:} LogicOCR comprises 2,780 questions, including two subsets, \ie, LogicOCR-Gen with 1,100 multi-choice questions on generated images, and LogicOCR-Real with 1,680 free-form question on real-world images.

\noindent \textbf{3. Does the dataset contain all possible instances or is it a sample (not necessarily random) of instances from a larger set? If the dataset is a sample, then what is the larger set? Is the sample representative of the larger set (e.g., geographic coverage)? If so, please describe how this representativeness was validated/verified. If it is not representative of the larger set, please describe why not (e.g., to cover a more diverse range of instances, because instances were withheld or unavailable).}

\textbf{A3:} In LogicOCR-Gen, the collected text corpora source from the Chinese National Civil Servant Examination, \ie, the larger set. Although the latest questions in  Chinese National Civil Servant Examination are not included, LogicOCR-Gen is still representative, covering categorical, sufficient conditional, necessary conditional, disjunctive, and conjunctive reasoning types. The images in LogicOCR-Real are collected from existing datasets and publicly available web sources as described in the main paper. The questions are manually crafted, featuring more intensive multi-hop reasoning.

\noindent \textbf{4. What data does each instance consist of? ``Raw'' data (e.g., unprocessed text or images)or features? In either case, please provide a description.}

\textbf{A4:} Each instance contains an image and an instruction. In LogicOCR-Gen, a multi-choice question is rendered on each image. The question corpora source from the Chinese National Civil Servant Examination. GPT-Image-1 is leveraged to generate diverse images based on the curated corpora. In LogicOCR-Real, the free-form questions are provided in text form.

\noindent \textbf{5. Is there a label or target associated with each instance? If so, please provide a description.}

\textbf{A5:} Yes, each instance contains exactly one target as the true answer corresponding to the question.

\noindent \textbf{6. Is any information missing from individual instances? If so, please provide a description, explaining why this information is missing (e.g., because it was unavailable). This does not include intentionally removed information, but might include, e.g., redacted text.}

\textbf{A6:} Very few images in LogicOCR-Gen miss the last option's content. The missing option is cut out by GPT-Image-1 during text-to-image generation. We manually verify that the missing option is not the correct answer and does not affect problem-solving.

\noindent \textbf{7. Are relationships between individual instances made explicit (e.g., users' movie ratings, social network links)? If so, please describe how these relationships are made explicit.}

\textbf{A7:} No, there is no explicit relationship between each instance.

\noindent \textbf{8. Are there recommended data splits (e.g., training, development/validation, testing)? If so, please provide a description of these splits, explaining the rationale behind them.}

\textbf{A8:} Yes, we recommend using all instances of LogicOCR for testing.

\noindent \textbf{9. Are there any errors, sources of noise, or redundancies in the dataset? If so, please provide a description.}

\textbf{A9:} The raw text corpora of LogicOCR-Gen are sourced from the Chinese National Civil Servant Examination and have been curated by LogiQA~\cite{logiqa} and LogiQA2.0~\cite{logiqa2.0}. We reuse the corpora from LogiQA and LogiQA2.0, then further conduct deduplication. There are no redundancies in LogicOCR.

\noindent \textbf{10. Is the dataset self-contained, or does it link to or otherwise rely on external resources (e.g., websites, tweets, other datasets)? If it links to or relies on external resources, a) are there guarantees that they will exist, and remain constant, over time; b) are there official archival versions of the complete dataset (i.e., including the external resources as they existed at the time the dataset was created); c) are there any restrictions (e.g., licenses, fees) associated with any of the external resources that might apply to a future user? Please provide descriptions of all external resources and any restrictions associated with them, as well as links or other access points, as appropriate.}

\textbf{A10:} The dataset is self-contained. The official archival versions of the complete dataset can be downloaded from GitHub. There are no fees. The data should be used in non-commercial scenarios following CC BY-NC-SA 4.0 license.

\noindent \textbf{11. Does the dataset contain data that might be considered confidential (e.g., data that is protected by legal privilege or by doctorpatient confidentiality, data that includes the content of individuals non-public communications)? If so, please provide a description.}

\textbf{A11:} No.

\noindent \textbf{12. Does the dataset contain data that, if viewed directly, might be offensive, insulting, threatening, or might otherwise cause anxiety? If so, please describe why.}

\textbf{A12:} No.

\subsection{Collection Process}

\noindent \textbf{1. How was the data associated with each instance acquired? Was the data directly observable (e.g., raw text, movie ratings), reported by subjects (e.g., survey responses), or indirectly inferred/derived from other data (e.g., part-of-speech tags, model-based guesses for age or language)? If data was reported by subjects or indirectly inferred/derived from other data, was the data validated/verified? If so, please describe how.}

\textbf{A1:} The data associated with each instance is directly observable.

\noindent \textbf{2. What mechanisms or procedures were used to collect the data (e.g., hardware apparatus or sensor, manual human curation, software program, software API)? How were these mechanisms or procedures validated?}

\textbf{A2:} For LogicOCR-Gen, the text corpora are reused from LogiQA and LogiQA2.0. GPT-Image-1 API is used from transfer the text corpora in to diverse images. The images are manually verified. For LogicOCR-Real, the images are manually selected and the questions are meticuously designed by experts.

\noindent \textbf{3. If the dataset is a sample from a larger set, what was the sampling strategy (e.g., deterministic, probabilistic with specific sampling probabilities)?}

\textbf{A3:} Cosine similarity of the sentence embeddings is used for text corpora deduplication. Then, the generated images based on the text corpora are filtered according to subjective quality assessment.

\noindent \textbf{4. Who was involved in the data collection process (e.g., students, crowdworkers, contractors) and how were they compensated (e.g., how much were crowdworkers paid)?}

\textbf{A4:} Only the authors of this paper.

\noindent \textbf{5. Over what timeframe was the data collected? Does this timeframe match the creation timeframe of the data associated with the instances (e.g., recent crawl of old news articles)? If not, please describe the timeframe in which the data associated with the instances was created.}

\textbf{A5}: The timeframe for the generated images is from April to May 2025.

\subsection{Preprocessing/cleaning/labeling}

\noindent \textbf{1. Was any preprocessing/cleaning/labeling of the data done (e.g., discretization or bucketing, tokenization, part-of-speech tagging, SIFT feature extraction, removal of instances, processing of missing values)? If so, please provide a description. If not, you may skip the remainder of the questions in this section.}

\textbf{A1:} For individual samples with mixed languages in the collected text corpora, we retranslate the original Chinese corpus into English with Qwen2.5~\cite{qwen2.5}. The text corpus of each sample can be divided into three parts, including \textit{context}, \textit{question}, and \textit{options}. The context part provides all the information required to solve the question. Some different questions share the same context corpus. We use the context part of each sample for further deduplication. We first convert each context corpus into one embedding using BGE-M3~\cite{chen2024bge} and then calculate the cosine similarity for deduplication. 

\noindent \textbf{2. Was the ``raw'' data saved in addition to the preprocessed/cleaned/labeled data (e.g., to support unanticipated future uses)? If so, please provide a link or other access point to the ``raw'' data.}

\textbf{A2:} For LogicOCR|-Gen, the raw text corpora can be found in \href{https://github.com/lgw863/LogiQA-dataset}{LogiQA} and \href{https://github.com/csitfun/LogiQA2.0}{LogiQA2.0}

\noindent \textbf{3. Is the software used to preprocess/clean/label the instances available? If so, please provide a link or other access point.}

\textbf{A3:} Link for \href{https://huggingface.co/Qwen/Qwen2.5-14B-Instruct}{Qwen2.5} and \href{https://huggingface.co/BAAI/bge-m3}{BGE-M3}.

\subsection{Uses}

\noindent \textbf{1. Has the dataset been used for any tasks already? If so, please provide a description.}

\textbf{A1:} No.

\noindent \textbf{2. Is there a repository that links to any or all papers or systems that use the dataset? If so, please provide a link or other access point.}

\textbf{A2:} No.

\noindent \textbf{3. What (other) tasks could the dataset be used for?}

\textbf{A3:} It can be used for testing and evaluating the LMMs' pure logical reasoning performance on text-rich images, while minimizing the reliance on STEM knowledge.

\noindent \textbf{4. Is there anything about the composition of the dataset or the way it was collected and preprocessed/cleaned/labeled that might impact future uses? For example, is there anything that a future user might need to know to avoid uses that could result in unfair treatment of individuals or groups (e.g., stereotyping, quality of service issues) or other undesirable harms (e.g., financial harms, legal risks) If so, please provide a description. Is there anything a future user could do to mitigate these undesirable harms?}

\textbf{A4:} No.

\noindent \textbf{5. Are there tasks for which the dataset should not be used? If so, please provide a description.}

\textbf{A5:} The dataset is a test set. It should not be used for training.

\subsection{Distribution}

\noindent \textbf{1. Will the dataset be distributed to third parties outside of the entity (e.g., company, institution, organization) on behalf of which the dataset was created? If so, please provide a description.}

\textbf{A1:} Yes.

\noindent \textbf{2. How will the dataset will be distributed (e.g., tarball on website, API, GitHub)? Does the dataset have a digital object identifier (DOI)?}

\textbf{A2:} It will be publicly available at \href{https://github.com/MiliLab/LogicOCR}{LogicOCR}.

\noindent \textbf{3. When will the dataset be distributed?}

\textbf{A3:} The dataset will be distributed in Nov. 2025.

\noindent \textbf{4. Will the dataset be distributed under a copyright or other intellectual property (IP) license, and/or under applicable terms of use (ToU)? If so, please describe this license and/or ToU, and provide a link or other access point to, or otherwise reproduce, any relevant licensing terms or ToU, as well as any fees associated with these restrictions.}

\textbf{A4:} It will be distributed under the \href{https://creativecommons.org/licenses/by-nc-sa/4.0/}{Creative Commons Attribution-NonCommercial-ShareAlike 4.0 License}.

\noindent \textbf{5. Have any third parties imposed IP-based or other restrictions on the data associated with the instances? If so, please describe these restrictions, and provide a link or other access point to, or otherwise reproduce, any relevant licensing terms, as well as any fees associated with these restrictions.}

\textbf{A5:} No.

\noindent \textbf{6. Do any export controls or other regulatory restrictions apply to the dataset or to individual instances? If so, please describe these restrictions, and provide a link or other access point to, or otherwise reproduce, any supporting documentation.}

\textbf{A6:} No.

\subsection{Maintenance}

\noindent \textbf{1. Who will be supporting/hosting/maintaining the dataset?}

\textbf{A1:} The authors.

\noindent \textbf{2. How can the owner/curator/manager of the dataset be contacted (e.g., email address)?}

\textbf{A2:} They can be contacted via email available on the paper.

\noindent \textbf{3. Is there an erratum? If so, please provide a link or other access point.}

\textbf{A3:} No. 

\noindent \textbf{4. Will the dataset be updated (e.g., to correct labeling errors, add new instances, delete instances)? If so, please describe how often, by whom, and how updates will be communicated to users (e.g., mailing list, GitHub)?}

\textbf{A4:} Yes. It will be updated by the authors of the paper and released on GitHub.

\noindent \textbf{5. Will older versions of the dataset continue to be supported/hosted/maintained? If so, please describe how. If not, please describe how its obsolescence will be communicated to users.}

\textbf{A5:} Yes. We will also provide the download link of the older versions.

\noindent \textbf{6. If others want to extend/augment/build on/contribute to the dataset, is there a mechanism for them to do so? If so, please provide a description. Will these contributions be validated/verified? If so, please describe how. If not, why not? Is there a process for communicating/distributing these contributions to other users? If so, please provide a description.}

\textbf{A6:} Others can contribute through GitHub. The contributions will be verified.

\newpage
\section{Case Study}


\begin{figure*}[!h]
    \centering
    \includegraphics[width=1\linewidth]{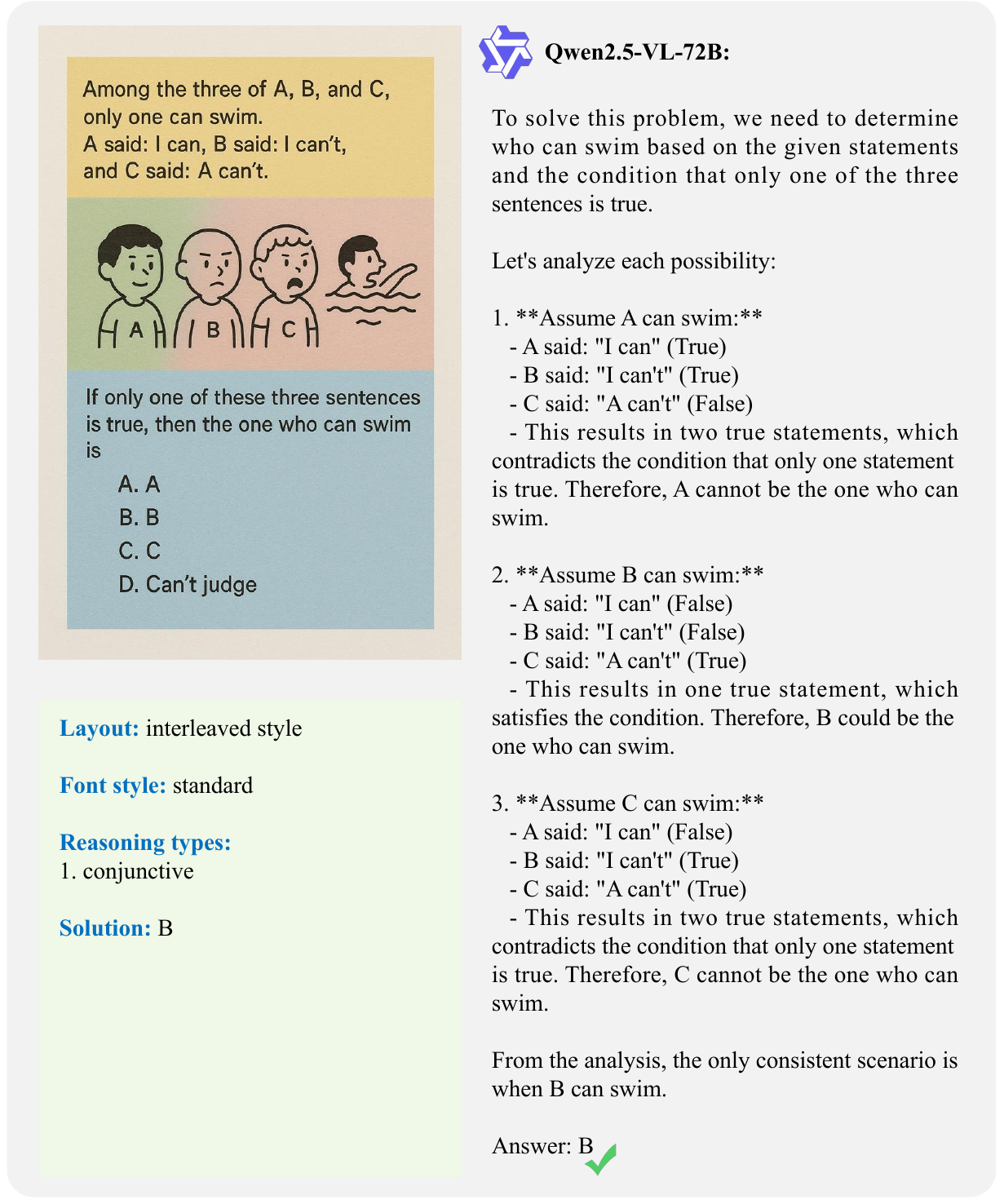}
    \caption{A correct case of Qwen2.5-VL-72B on the question involving one reasoning type from LogicOCR-Gen.}
\end{figure*}

\begin{figure*}[!h]
    \centering
    \includegraphics[width=1\linewidth]{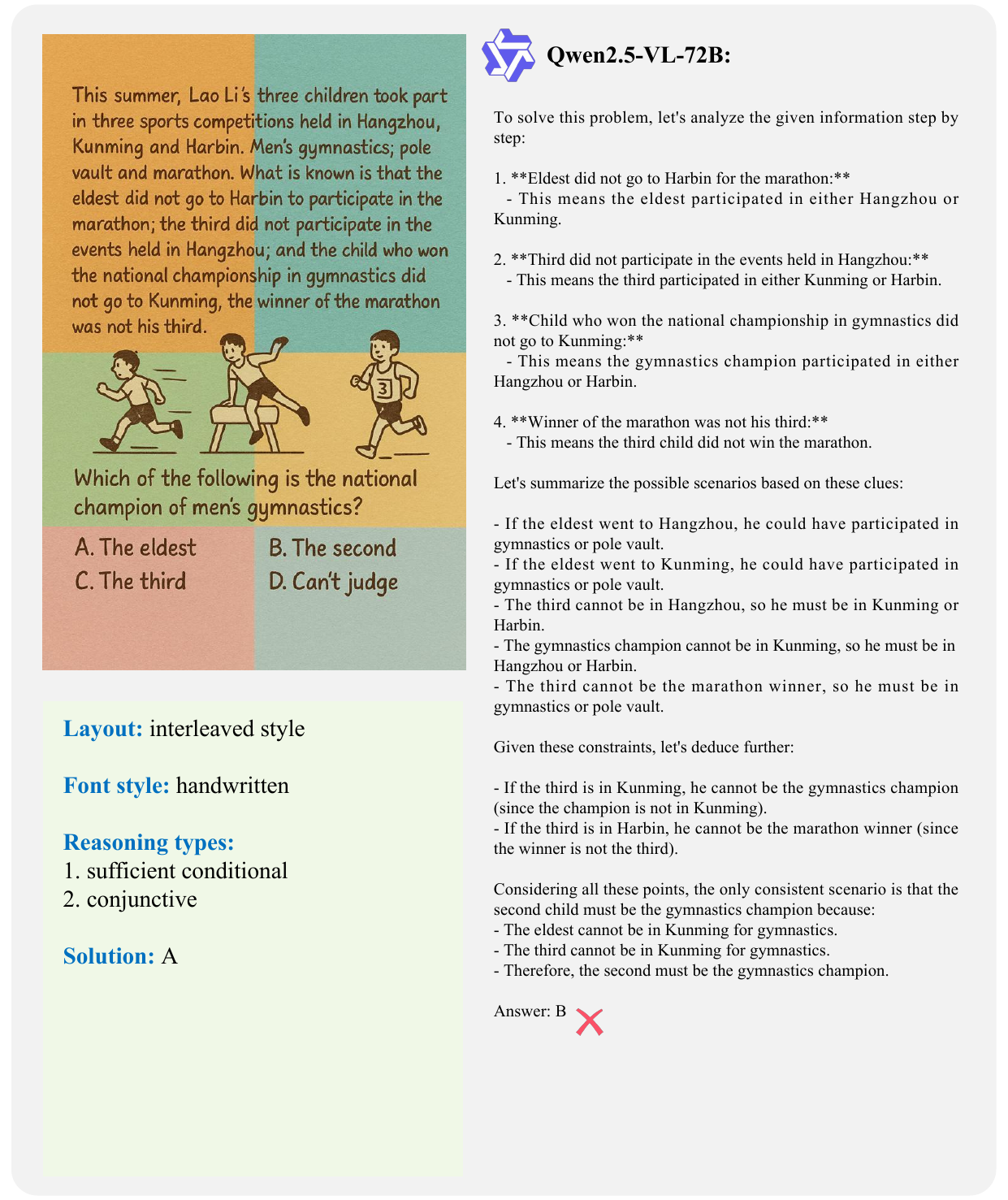}
    \caption{A failure case of Qwen2.5-VL-72B on the question involving two reasoning types from LogicOCR-Gen.}
\end{figure*}

\begin{figure*}[!h]
    \centering
    \includegraphics[width=1\linewidth]{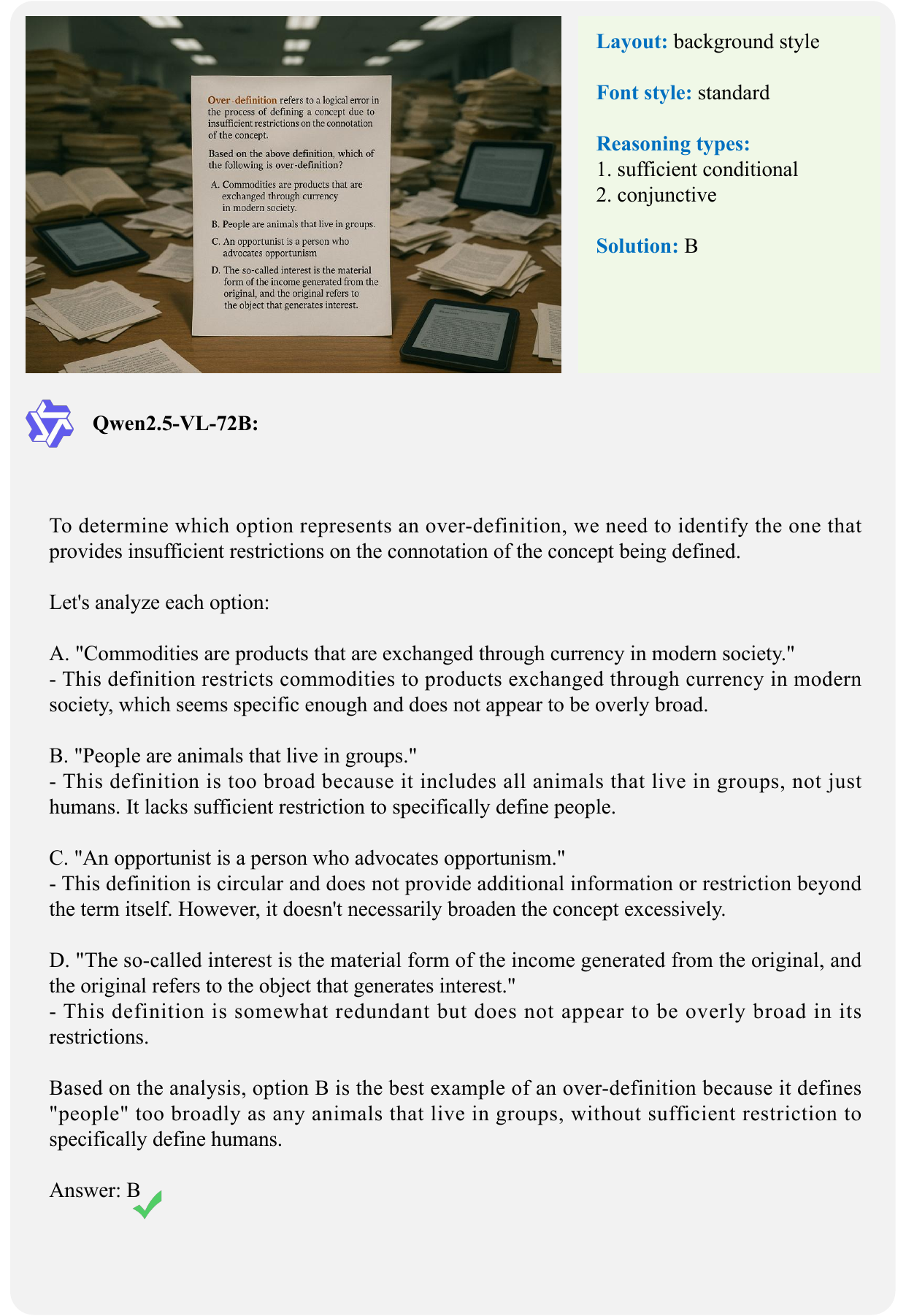}
    \caption{A correct case of Qwen2.5-VL-72B on the question involving two reasoning types from LogicOCR-Gen.}
\end{figure*}

\begin{figure*}[!h]
    \centering
    \includegraphics[width=1\linewidth]{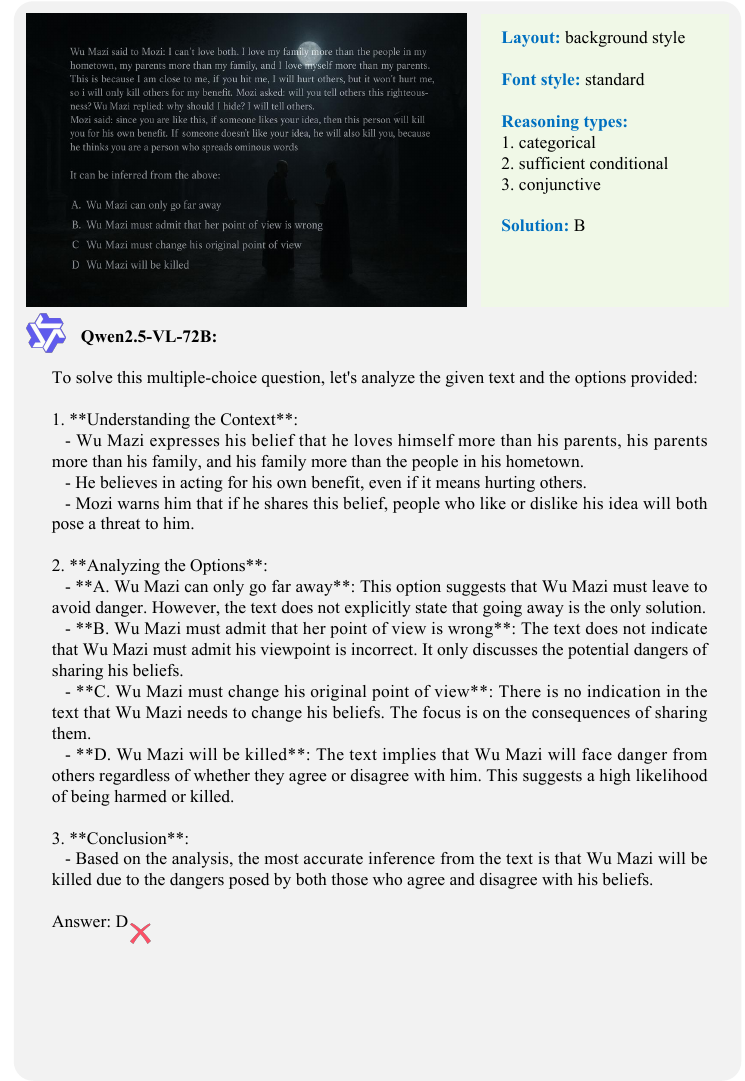}
    \caption{A failure case of Qwen2.5-VL-72B on the question involving three reasoning types from LogicOCR-Gen.}
\end{figure*}


\begin{figure*}[!h]
    \centering
    \includegraphics[width=1\linewidth]{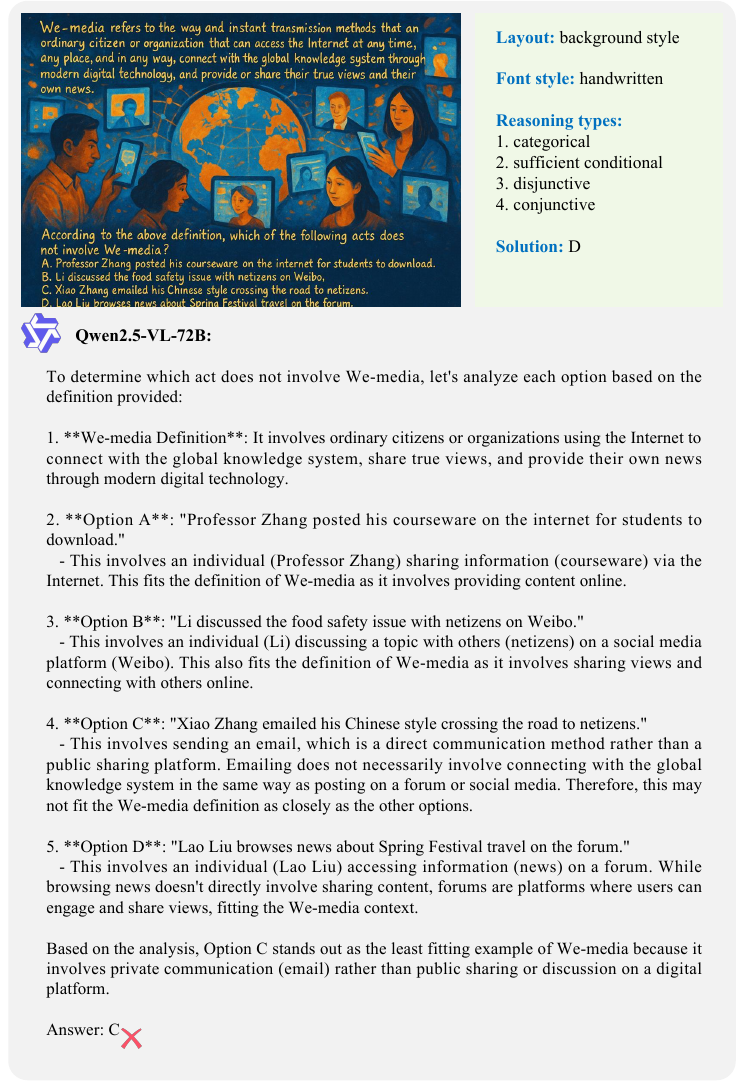}
    \caption{A failure case of Qwen2.5-VL-72B on the question involving four reasoning types from LogicOCR-Gen.}
\end{figure*}


\begin{figure*}[!h]
    \centering
    \includegraphics[width=1\linewidth]{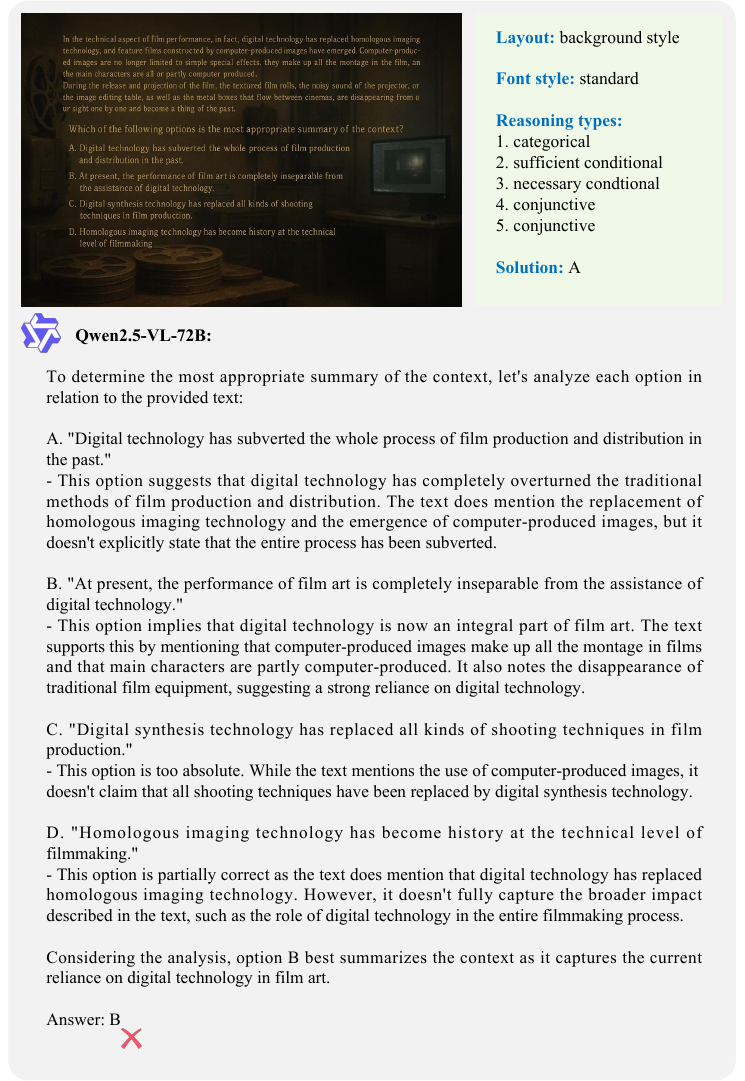}
    \caption{A failure case of Qwen2.5-VL-72B on the question involving five reasoning types from LogicOCR-Gen.}
\end{figure*}

\begin{figure*}[!h]
    \centering
    \includegraphics[width=1\linewidth]{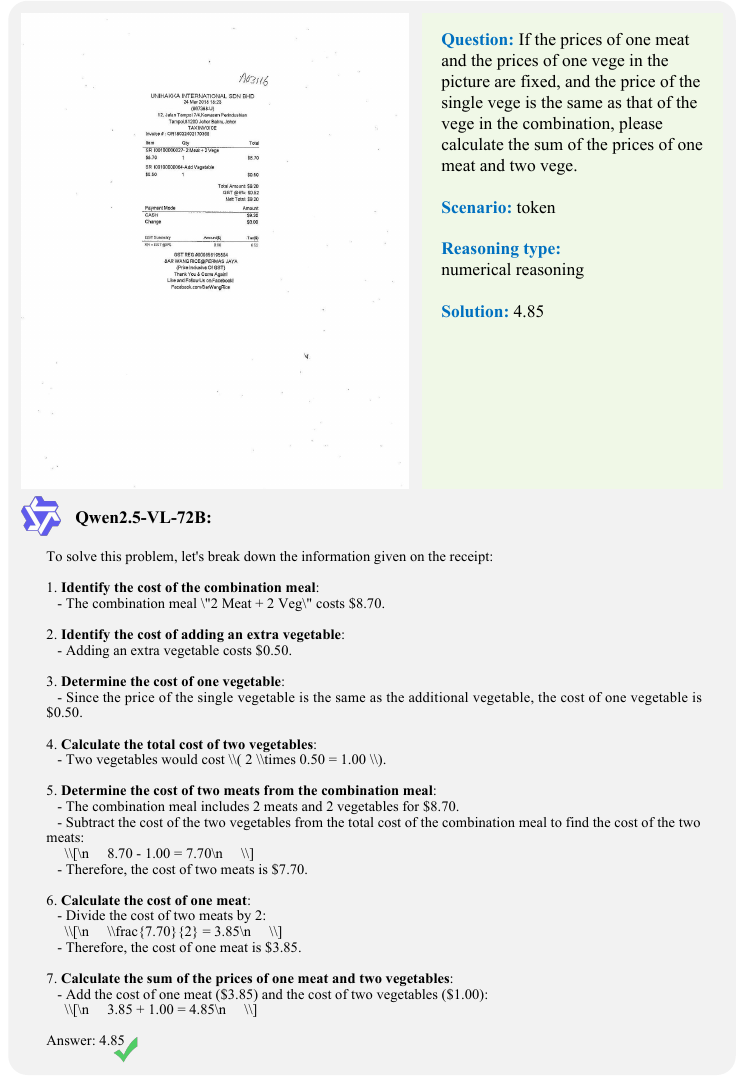}
    \caption{A correct case of Qwen2.5-VL-72B on the numerical reasoning question from LogicOCR-Real.}
\end{figure*}
\begin{figure*}[!h]
    \centering
    \includegraphics[width=1\linewidth]{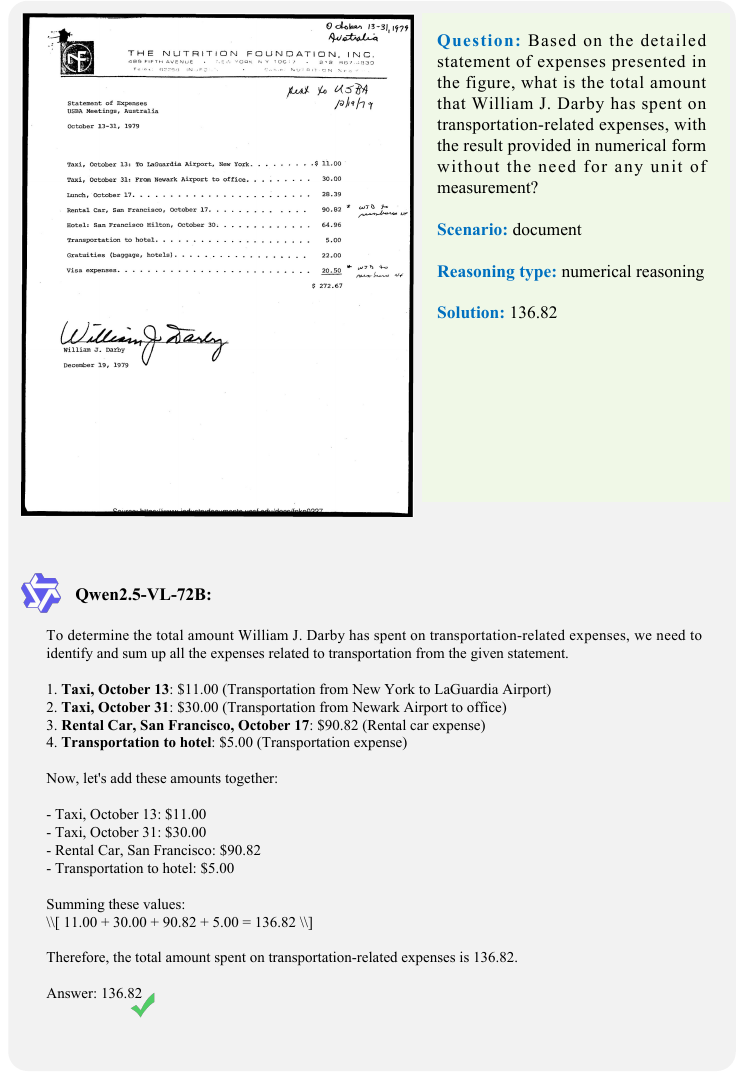}
    \caption{A correct case of Qwen2.5-VL-72B on the numerical reasoning question from LogicOCR-Real.}
\end{figure*}
\begin{figure*}[!h]
    \centering
    \includegraphics[width=1\linewidth]{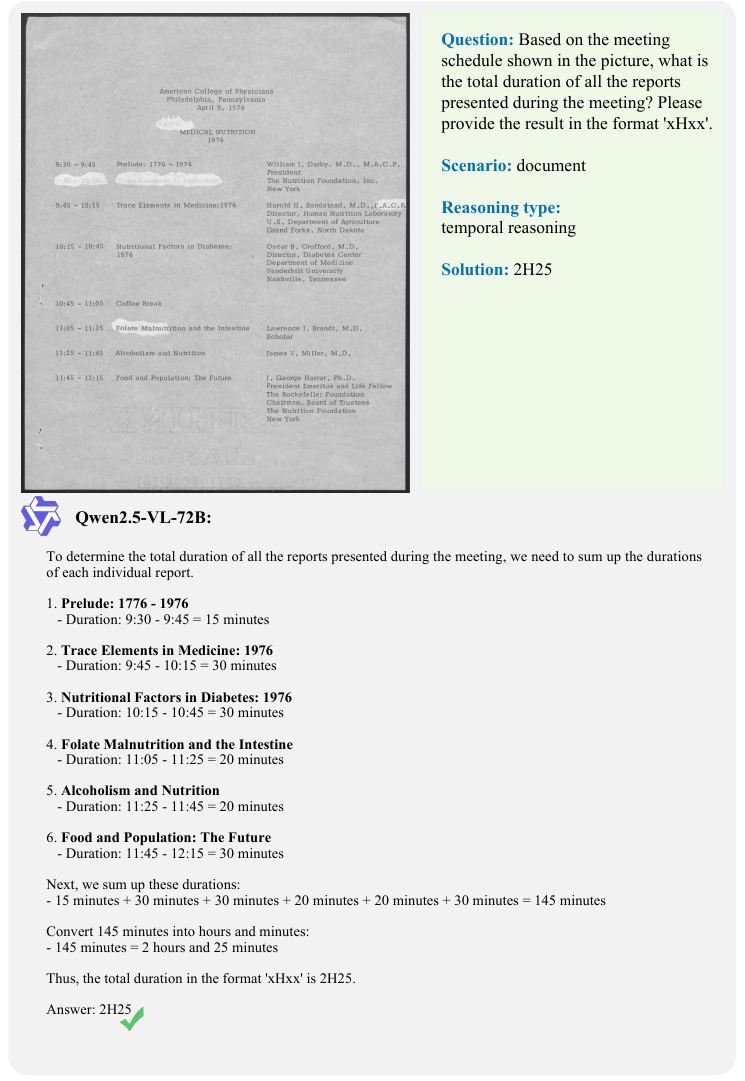}
    \caption{A correct case of Qwen2.5-VL-72B on the temporal reasoning question from LogicOCR-Real.}
\end{figure*}
\begin{figure*}[!h]
    \centering
    \includegraphics[width=1\linewidth]{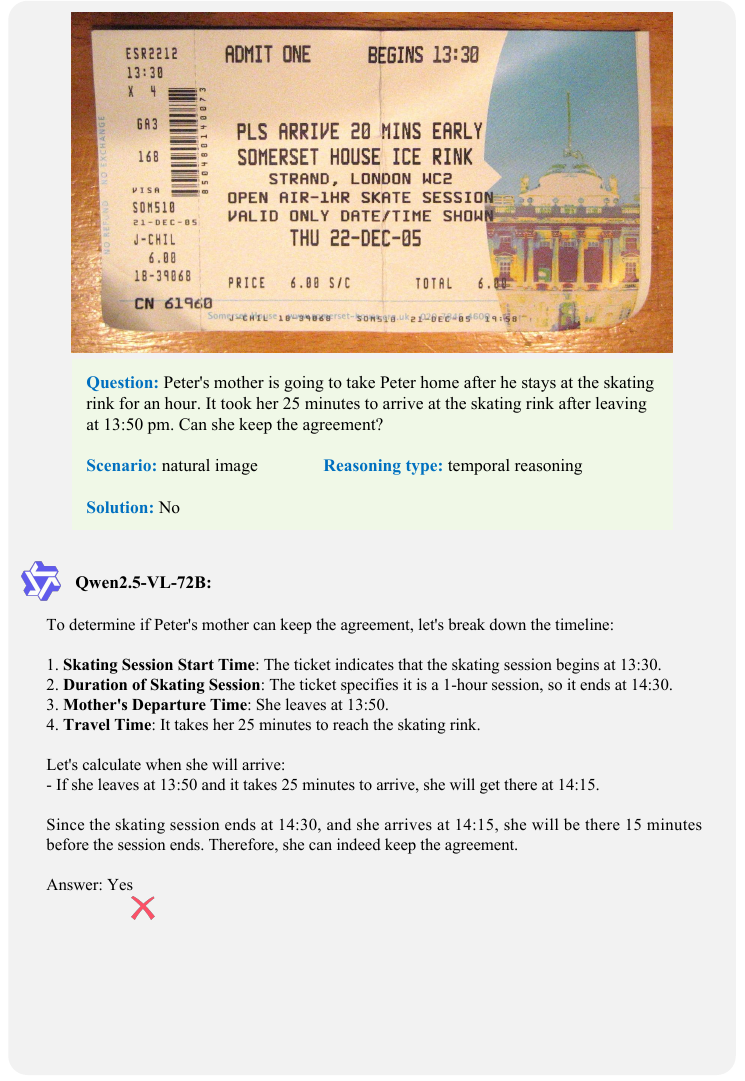}
    \caption{A failure case of Qwen2.5-VL-72B on the temporal reasoning question from LogicOCR-Real.}
\end{figure*}
\begin{figure*}[!h]
    \centering
    \includegraphics[width=1\linewidth]{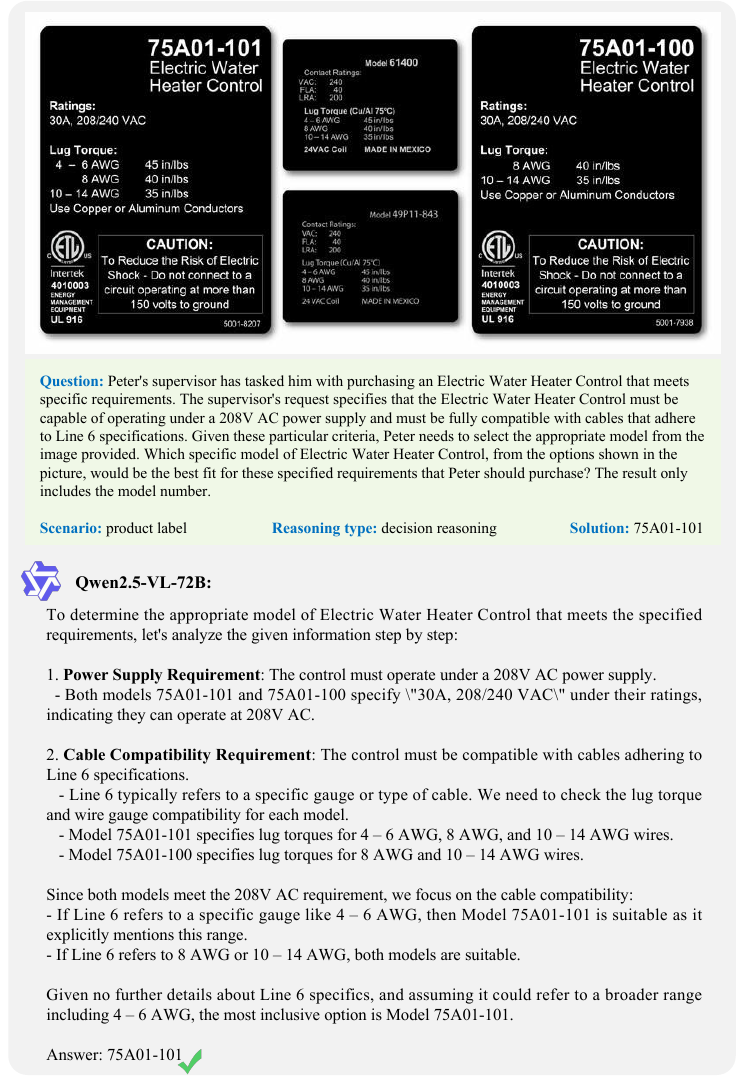}
    \caption{A correct case of Qwen2.5-VL-72B on the decision reasoning question from LogicOCR-Real.}
\end{figure*}
\begin{figure*}[!h]
    \centering
    \includegraphics[width=1\linewidth]{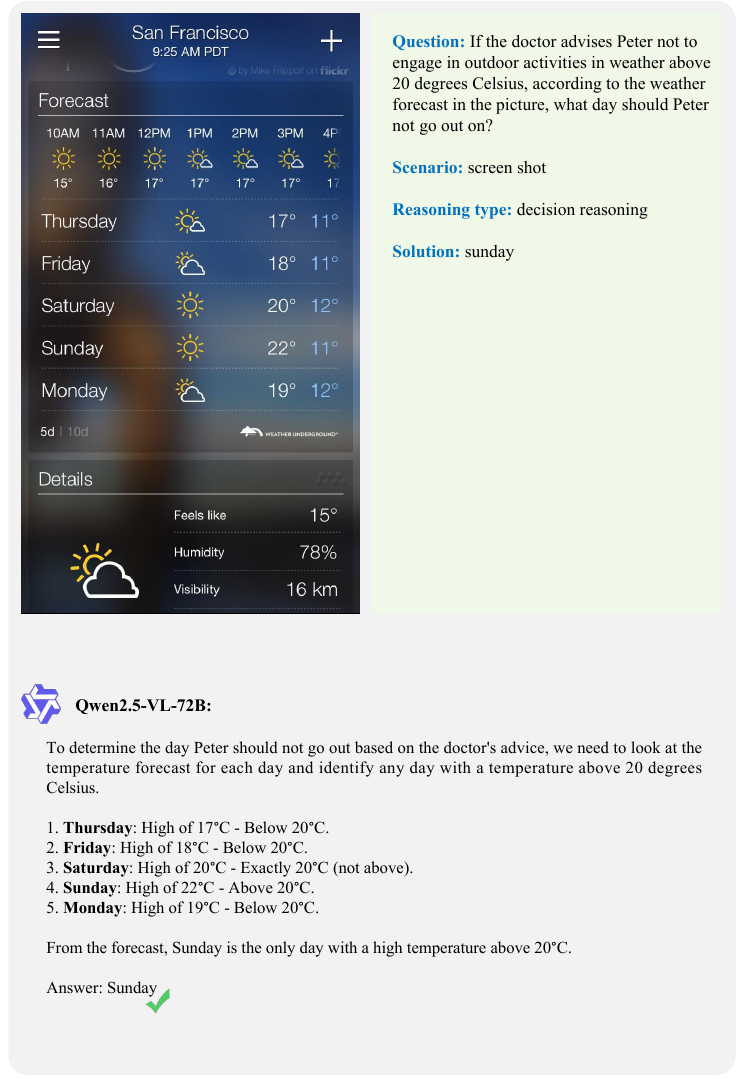}
    \caption{A correct case of Qwen2.5-VL-72B on the decision reasoning question from LogicOCR-Real.}
\end{figure*}
\begin{figure*}[!h]
    \centering
    \includegraphics[width=1\linewidth]{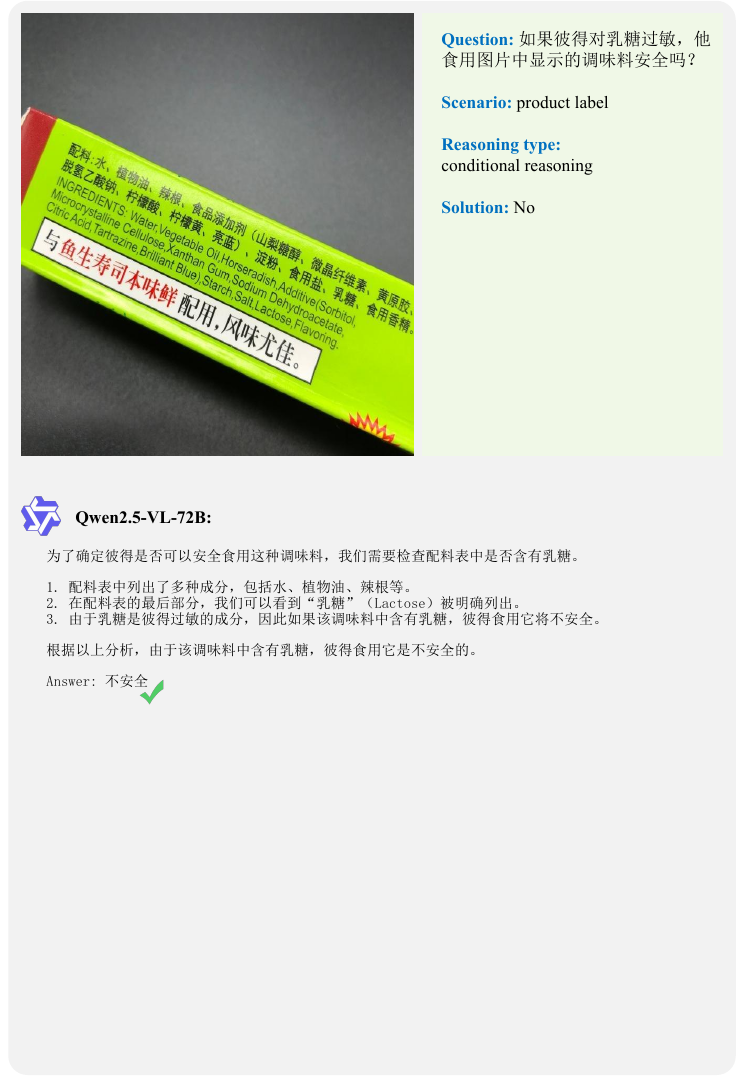}
    \caption{A correct case of Qwen2.5-VL-72B on the conditional reasoning question from LogicOCR-Real.}
\end{figure*}
\begin{figure*}[!h]
    \centering
    \includegraphics[width=1\linewidth]{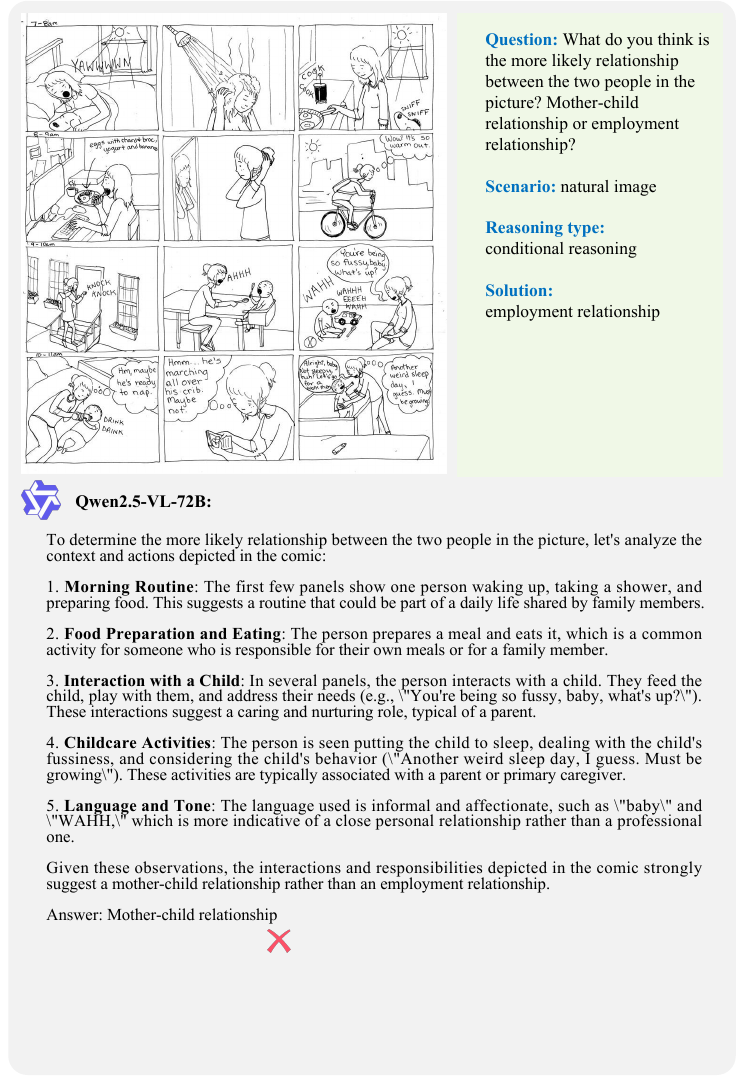}
    \caption{A failure case of Qwen2.5-VL-72B on the conditional reasoning question from LogicOCR-Real.}
\end{figure*}


\begin{figure*}[!h]
    \centering
    \includegraphics[width=1\linewidth]{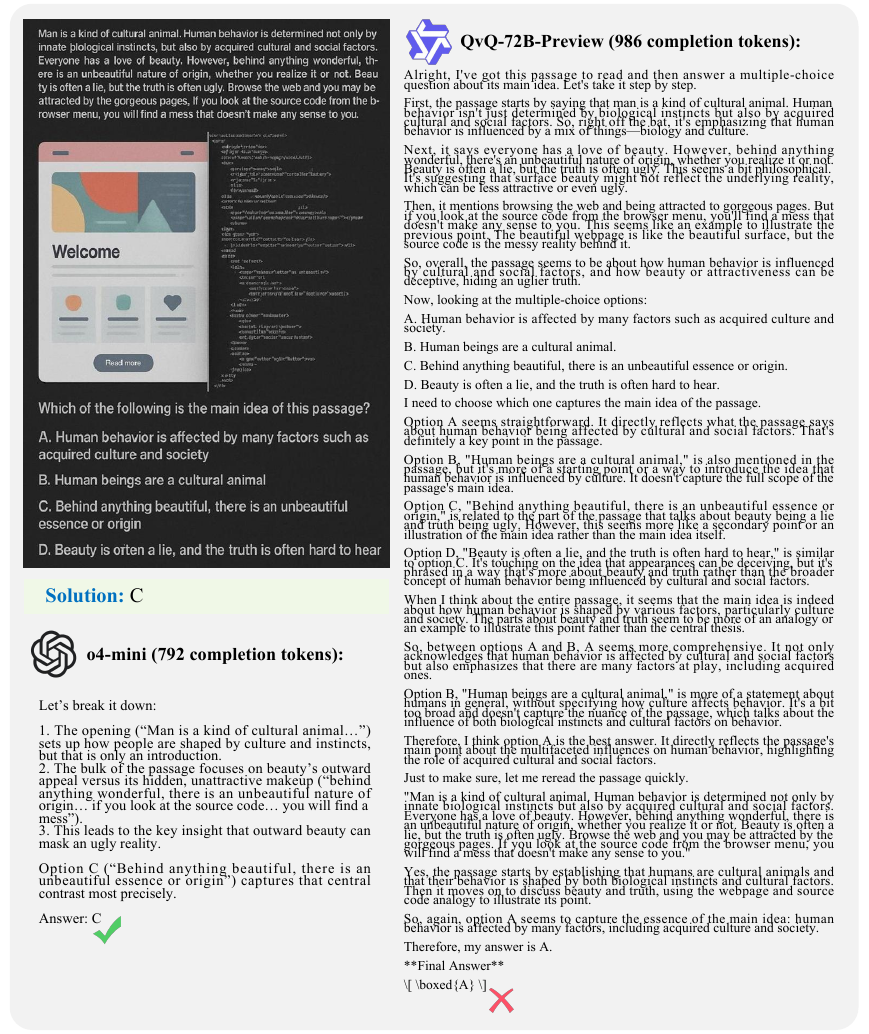}
    \caption{The response comparison of o4-mini and QvQ-72B-Preview (the thinking process of o4-mini cannot be achieved by user). In this case, o4-mini got correct answer with fewer output length.}
\end{figure*}

\begin{figure*}[!h]
    \centering
    \includegraphics[width=1\linewidth]{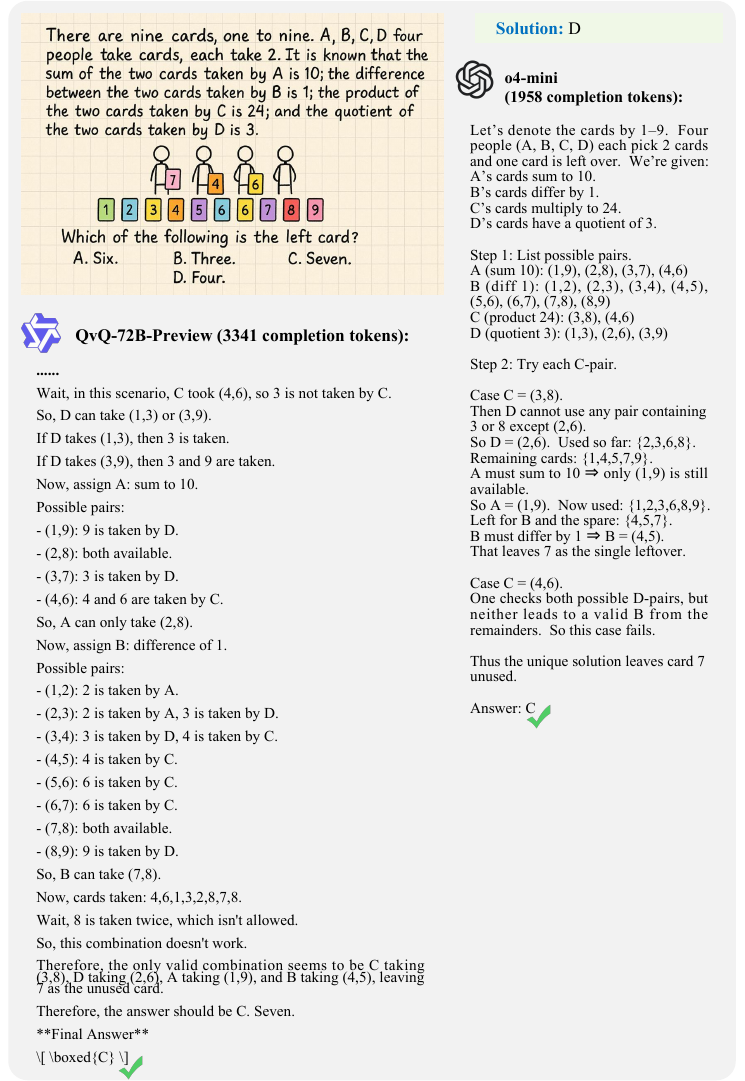}
    \caption{The response comparison of o4-mini and QvQ-72B-Preview. In this case, compared to QvQ-72B-Preview, o4-mini only uses 58.6\% completion tokens to obtain the correct choice.}
\end{figure*}

\clearpage

\bibliographystyle{elsarticle-num}
\bibliography{citation}

@article{kimivl,
  title={Kimi-vl technical report},
  author={Team, Kimi and Du, Angang and Yin, Bohong and Xing, Bowei and Qu, Bowen and Wang, Bowen and Chen, Cheng and Zhang, Chenlin and Du, Chenzhuang and Wei, Chu and others},
  journal={arXiv preprint arXiv:2504.07491},
  year={2025}
}

@article{liu2024ocrbench,
  title={OCRBench: on the hidden mystery of OCR in large multimodal models},
  author={Liu, Yuliang and Li, Zhang and Huang, Mingxin and Yang, Biao and Yu, Wenwen and others},
  journal={Science China Information Sciences},
  volume={67},
  number={12},
  pages={220102},
  year={2024},
}

@article{internvl3,
  title={InternVL3: Exploring Advanced Training and Test-Time Recipes for Open-Source Multimodal Models},
  author={Zhu, Jinguo and Wang, Weiyun and Chen, Zhe and Liu, Zhaoyang and Ye, Shenglong and Gu, Lixin and Duan, Yuchen and Tian, Hao and Su, Weijie and Shao, Jie and others},
  journal={arXiv preprint arXiv:2504.10479},
  year={2025}
}

@article{qwen2_5vl,
  title={Qwen2. 5-vl technical report},
  author={Bai, Shuai and Chen, Keqin and Liu, Xuejing and Wang, Jialin and Ge, Wenbin and Song, Sibo and Dang, Kai and Wang, Peng and Wang, Shijie and Tang, Jun and others},
  journal={arXiv preprint arXiv:2502.13923},
  year={2025}
}

@article{liu2024textmonkey,
  title={Textmonkey: An ocr-free large multimodal model for understanding document},
  author={Liu, Yuliang and Yang, Biao and Liu, Qiang and Li, Zhang and others},
  journal={arXiv preprint arXiv:2403.04473},
  year={2024}
}

@article{mplug-docowl2,
  title={mplug-docowl2: High-resolution compressing for ocr-free multi-page document understanding},
  author={Hu, Anwen and Xu, Haiyang and Zhang, Liang and Ye, Jiabo and Yan, Ming and others},
  journal={arXiv preprint arXiv:2409.03420},
  year={2024}
}

@misc{qvq,
    title = {QVQ: To See the World with Wisdom},
    url = {https://qwenlm.github.io/blog/qvq-72b-preview/},
    author = {Qwen Team},
    month = {December},
    year = {2024}
}

@inproceedings{singh2019towards,
  title={Towards vqa models that can read},
  author={Singh, Amanpreet and Natarajan, Vivek and Shah, Meet and Jiang, Yu and Chen, Xinlei and others},
  booktitle={CVPR},
  pages={8317--8326},
  year={2019}
}

@inproceedings{mathew2021docvqa,
  title={Docvqa: A dataset for vqa on document images},
  author={Mathew, Minesh and Karatzas, Dimosthenis and Jawahar, CV},
  booktitle={WACV},
  pages={2200--2209},
  year={2021}
}

@inproceedings{masry2022chartqa,
  title={ChartQA: A Benchmark for Question Answering about Charts with Visual and Logical Reasoning},
  author={Masry, Ahmed and Do, Xuan Long and Tan, Jia Qing and Joty, Shafiq and Hoque, Enamul},
  booktitle={Findings of ACL},
  pages={2263--2279},
  year={2022}
}

@inproceedings{dtvqa,
  title={Exploring the capabilities of large multimodal models on dense text},
  author={Zhang, Shuo and Yang, Biao and Li, Zhang and Ma, Zhiyin and Liu, Yuliang and Bai, Xiang},
  booktitle={ICDAR},
  pages={281--298},
  year={2024},
}

@inproceedings{ocrbenchv2,
  title={OCRBench v2: An Improved Benchmark for Evaluating Large Multimodal Models on Visual Text Localization and Reasoning},
  author={Fu, Ling and Yang, Biao and Kuang, Zhebin and Song, Jiajun and Li, Yuzhe and others},
  booktitle={NeurIPS},
  year={2025}
}

@article{ccocr,
  title={CC-OCR: A Comprehensive and Challenging OCR Benchmark for Evaluating Large Multimodal Models in Literacy},
  author={Yang, Zhibo and Tang, Jun and Li, Zhaohai and Wang, Pengfei and Wan, Jianqiang and others},
  journal={arXiv preprint arXiv:2412.02210},
  year={2024}
}

@inproceedings{lumathvista,
  title={MathVista: Evaluating Mathematical Reasoning of Foundation Models in Visual Contexts},
  author={Lu, Pan and Bansal, Hritik and Xia, Tony and Liu, Jiacheng and Li, Chunyuan and others},
  booktitle={ICLR},
  year={2024}
}

@inproceedings{zhang2025mathverse,
  title={Mathverse: Does your multi-modal llm truly see the diagrams in visual math problems?},
  author={Zhang, Renrui and Jiang, Dongzhi and Zhang, Yichi and Lin, Haokun and Guo, Ziyu and others},
  booktitle={ECCV},
  pages={169--186},
  year={2025},
}

@misc{gpt4o,
  author       = {OpenAI},
  title        = {{Hello GPT-4o}},
  year         = {2024},
  howpublished = {\url{https://openai.com/index/gpt-4v-system-card}},
  note         = {Accessed: 2024-12-29}
}

@misc{o1,
  author       = {OpenAI},
  title        = {{OpenAI o1 System Card}},
  year         = {2024},
  howpublished = {\url{https://openai.com/index/openai-o1-system-card/}},
  note         = {Accessed: 2024-12-05}
}

@misc{o4mini,
  author       = {OpenAI},
  title        = {{OpenAI o3 and o4-mini System Card}},
  year         = {2025},
  howpublished = {\url{https://openai.com/index/o3-o4-mini-system-card/}},
  note         = {Accessed: 2025-04-16}
}

@misc{gptimage1,
  author       = {OpenAI},
  title        = {{Introducing our latest image generation model in the API}},
  year         = {2025},
  howpublished = {\url{https://openai.com/index/image-generation-api/}},
  note         = {Accessed: 2025-04-23}
}

@misc{gemini25pro,
  author       = {DeepMind},
  title        = {{Gemini 2.5 Pro}},
  year         = {2025},
  howpublished = {\url{https://deepmind.google/technologies/gemini/pro/}},
}

@misc{claude3.7sonnet,
  author       = {Anthropic},
  title        = {{Claude 3.7 Sonnet and Claude Code}},
  year         = {2025},
  howpublished = {\url{https://www.anthropic.com/news/claude-3-7-sonnet}},
  note         = {Accessed: 2025-02-25}
}

@article{deepseekvl2,
  title={Deepseek-vl2: Mixture-of-experts vision-language models for advanced multimodal understanding},
  author={Wu, Zhiyu and Chen, Xiaokang and Pan, Zizheng and Liu, Xingchao and Liu, Wen and Dai, Damai and Gao, Huazuo and Ma, Yiyang and Wu, Chengyue and Wang, Bingxuan and others},
  journal={arXiv preprint arXiv:2412.10302},
  year={2024}
}

@article{nvila,
  title={NVILA: Efficient frontier visual language models},
  author={Liu, Zhijian and Zhu, Ligeng and Shi, Baifeng and Zhang, Zhuoyang and Lou, Yuming and Yang, Shang and Xi, Haocheng and Cao, Shiyi and Gu, Yuxian and Li, Dacheng and others},
  journal={arXiv preprint arXiv:2412.04468},
  year={2024}
}

@article{song2025visualpuzzles,
  title={VisualPuzzles: Decoupling Multimodal Reasoning Evaluation from Domain Knowledge},
  author={Song, Yueqi and Ou, Tianyue and Kong, Yibo and Li, Zecheng and Neubig, Graham and Yue, Xiang},
  journal={arXiv preprint arXiv:2504.10342},
  year={2025}
}

@inproceedings{mmmu,
  title={Mmmu: A massive multi-discipline multimodal understanding and reasoning benchmark for expert agi},
  author={Yue, Xiang and Ni, Yuansheng and Zhang, Kai and Zheng, Tianyu and Liu, Ruoqi and Zhang, Ge and Stevens, Samuel and Jiang, Dongfu and Ren, Weiming and Sun, Yuxuan and others},
  booktitle={CVPR},
  pages={9556--9567},
  year={2024}
}

@article{wang2024measuring,
  title={Measuring multimodal mathematical reasoning with math-vision dataset},
  author={Wang, Ke and Pan, Junting and Shi, Weikang and Lu, Zimu and Ren, Houxing and Zhou, Aojun and Zhan, Mingjie and Li, Hongsheng},
  journal={NeurIPS},
  volume={37},
  pages={95095--95169},
  year={2024}
}

@article{wangcharxiv,
  title={Charxiv: Charting gaps in realistic chart understanding in multimodal llms},
  author={Wang, Zirui and Xia, Mengzhou and He, Luxi and Chen, Howard and Liu, Yitao and Zhu, Richard and Liang, Kaiqu and Wu, Xindi and Liu, Haotian and Malladi, Sadhika and others},
  journal={NeurIPS},
  volume={37},
  pages={113569--113697},
  year={2024}
}

@article{mmmu-pro,
  title={Mmmu-pro: A more robust multi-discipline multimodal understanding benchmark},
  author={Yue, Xiang and Zheng, Tianyu and Ni, Yuansheng and Wang, Yubo and Zhang, Kai and Tong, Shengbang and Sun, Yuxuan and Yu, Botao and Zhang, Ge and Sun, Huan and others},
  journal={arXiv preprint arXiv:2409.02813},
  year={2024}
}

@inproceedings{logiqa,
  title={Logiqa: A challenge dataset for machine reading comprehension with logical reasoning},
  author={Liu, Jian and Cui, Leyang and Liu, Hanmeng and Huang, Dandan and Wang, Yile and Zhang, Yue},
  booktitle={IJCAI},
  year={2020}
}

@article{logiqa2.0,
  title={Logiqa 2.0—an improved dataset for logical reasoning in natural language understanding},
  author={Liu, Hanmeng and Liu, Jian and Cui, Leyang and Teng, Zhiyang and Duan, Nan and Zhou, Ming and Zhang, Yue},
  journal={IEEE/ACM Transactions on Audio, Speech, and Language Processing},
  volume={31},
  pages={2947--2962},
  year={2023}
}

@article{chen2024bge,
  title={Bge m3-embedding: Multi-lingual, multi-functionality, multi-granularity text embeddings through self-knowledge distillation},
  author={Chen, Jianlv and Xiao, Shitao and Zhang, Peitian and Luo, Kun and Lian, Defu and Liu, Zheng},
  journal={arXiv preprint arXiv:2402.03216},
  year={2024}
}

@article{qwen2.5,
  title={Qwen2. 5 technical report},
  author={Yang, An and Yang, Baosong and Zhang, Beichen and Hui, Binyuan and Zheng, Bo and Yu, Bowen and Li, Chengyuan and Liu, Dayiheng and Huang, Fei and Wei, Haoran and others},
  journal={arXiv preprint arXiv:2412.15115},
  year={2024}
}

@inproceedings{gupta2016synthetic,
  title={Synthetic data for text localisation in natural images},
  author={Gupta, Ankush and Vedaldi, Andrea and Zisserman, Andrew},
  booktitle={CVPR},
  pages={2315--2324},
  year={2016}
}

@article{ovis,
  title={Ovis: Structural Embedding Alignment for Multimodal Large Language Model},
  author={Shiyin Lu and Yang Li and Qing-Guo Chen and Zhao Xu and Weihua Luo and Kaifu Zhang and Han-Jia Ye},
  year={2024},
  journal={arXiv:2405.20797}
}

@article{sprague2024cot,
  title={To cot or not to cot? chain-of-thought helps mainly on math and symbolic reasoning},
  author={Sprague, Zayne and Yin, Fangcong and Rodriguez, Juan Diego and Jiang, Dongwei and Wadhwa, Manya and Singhal, Prasann and Zhao, Xinyu and Ye, Xi and Mahowald, Kyle and Durrett, Greg},
  journal={arXiv preprint arXiv:2409.12183},
  year={2024}
}

@article{guan2025token,
  title={A Token-level Text Image Foundation Model for Document Understanding},
  author={Guan, Tongkun and Wang, Zining and Fu, Pei and Guo, Zhengtao and Shen, Wei and Zhou, Kai and Yue, Tiezhu and Duan, Chen and Sun, Hao and Jiang, Qianyi and others},
  journal={arXiv preprint arXiv:2503.02304},
  year={2025}
}

@article{seed1.5-vl,
  title={Seed1.5-VL Technical Report
},
  author={Guo, Dong and Wu, Faming and Zhu, Feida and Leng, Fuxing and Shi, Guang and Chen, Haobin and Fan, Haoqi and others},
  journal={arXiv preprint arXiv:2505.07062},
  year={2025}
}

@inproceedings{mllm_know,
  title={MLLMs Know Where to Look: Training-free Perception of Small Visual Details with Multimodal LLMs},
  author={Zhang, Jiarui and Khayatkhoei, Mahyar and Chhikara, Prateek and Ilievski, Filip},
  booktitle={ICLR},
  year={2025},
}

@inproceedings{kangsee,
  title={See What You Are Told: Visual Attention Sink in Large Multimodal Models},
  author={Kang, Seil and Kim, Jinyeong and Kim, Junhyeok and Hwang, Seong Jae},
  booktitle={ICLR},
  year={2025},
}

@inproceedings{wu2024v,
  title={V?: Guided visual search as a core mechanism in multimodal llms},
  author={Wu, Penghao and Xie, Saining},
  booktitle={CVPR},
  year={2024}
}

@inproceedings{deka2017rico,
  title={Rico: A mobile app dataset for building data-driven design applications},
  author={Deka, Biplab and Huang, Zifeng and Franzen, Chad and Hibschman, Joshua and Afergan, Daniel and Li, Yang and Nichols, Jeffrey and Kumar, Ranjitha},
  booktitle={Proceedings of the 30th annual ACM symposium on user interface software and technology},
  pages={845--854},
  year={2017}
}

@article{wildreceipt,
  title={Spatial dual-modality graph reasoning for key information extraction},
  author={Sun, Hongbin and Kuang, Zhanghui and Yue, Xiaoyu and Lin, Chenhao and Zhang, Wayne},
  journal={arXiv preprint arXiv:2103.14470},
  year={2021}
}

@inproceedings{jaume2019funsd,
title={Funsd: A dataset for form understanding in noisy scanned documents},
author={Jaume, Guillaume and Ekenel, Hazim Kemal and Thiran, Jean-Philippe},
booktitle={2019 International Conference on Document Analysis and Recognition Workshops (ICDARW)},
volume={2},
pages={1--6},
year={2019}
}

@inproceedings{methani2020plotqa,
  title={Plotqa: Reasoning over scientific plots},
  author={Methani, Nitesh and Ganguly, Pritha and Khapra, Mitesh M and Kumar, Pratyush},
  booktitle={Proceedings of the ieee/cvf winter conference on applications of computer vision},
  pages={1527--1536},
  year={2020}
}

@inproceedings{he2024cmmu,
  title={CMMU: a benchmark for chinese multi-modal multi-type question understanding and reasoning},
  author={He, Zheqi and Wu, Xinya and Zhou, Pengfei and Xuan, Richeng and Liu, Guang and Yang, Xi and Zhu, Qiannan and Huang, Hua},
  booktitle={Proceedings of the Thirty-Third International Joint Conference on Artificial Intelligence},
  pages={830--838},
  year={2024}
}

@inproceedings{pu2025judge,
  title={Judge anything: Mllm as a judge across any modality},
  author={Pu, Shu and Wang, Yaochen and Chen, Dongping and Chen, Yuhang and Wang, Guohao and Qin, Qi and Zhang, Zhongyi and Zhang, Zhiyuan and Zhou, Zetong and Gong, Shuang and others},
  booktitle={Proceedings of the 31st ACM SIGKDD Conference on Knowledge Discovery and Data Mining V. 2},
  pages={5742--5753},
  year={2025}
}

@article{hisam,
  title={Hi-SAM: Marrying Segment Anything Model for Hierarchical Text Segmentation},
  author={Ye, Maoyuan and Zhang, Jing and Liu, Juhua and Liu, Chenyu and Yin, Baocai and Liu, Cong and Du, Bo and Tao, Dacheng},
  journal={IEEE Transactions on Pattern Analysis and Machine Intelligence},
  volume={47},
  number={3},
  pages={1431--1447},
  year={2025}
}

@article{llavaov1_5,
  title={Llava-onevision-1.5: Fully open framework for democratized multimodal training},
  author={An, Xiang and Xie, Yin and Yang, Kaicheng and Zhang, Wenkang and Zhao, Xiuwei and Cheng, Zheng and Wang, Yirui and Xu, Songcen and Chen, Changrui and Wu, Chunsheng and others},
  journal={arXiv preprint arXiv:2509.23661},
  year={2025}
}

@article{lu2025ovis2_5,
  title={Ovis2. 5 technical report},
  author={Lu, Shiyin and Li, Yang and Xia, Yu and Hu, Yuwei and Zhao, Shanshan and Ma, Yanqing and Wei, Zhichao and Li, Yinglun and Duan, Lunhao and Zhao, Jianshan and others},
  journal={arXiv preprint arXiv:2508.11737},
  year={2025}
}

\end{document}